%% file: arxiv_202601.tex
\documentclass[11pt]{article}

\usepackage[preprint]{acl}

\usepackage{times}
\usepackage{latexsym}
\usepackage{adjustbox}

\usepackage[T1]{fontenc}

\usepackage[utf8]{inputenc}

\usepackage{microtype}

\usepackage{inconsolata}

\usepackage{graphicx}

\usepackage{amsmath}
\usepackage{algorithm}
\usepackage{algpseudocode}

\usepackage{xspace}
\usepackage{booktabs}
\usepackage{multirow}
\usepackage{array}
\usepackage{subcaption}
\usepackage{xcolor}

\usepackage[breakable]{tcolorbox} 
\usepackage{inconsolata}        

\newcommand{\approach}{{ASAP}\xspace}
\newcommand{\model}{{ASAP}\xspace}

\title{Pruning the Unsurprising: Efficient LLM Reasoning via First-Token Surprisal}

\author{
  Wenhao Zeng\textsuperscript{1} \quad
  Yaoning Wang\textsuperscript{2} \quad
  Chao Hu\textsuperscript{1} \quad
  Yuling Shi\textsuperscript{1} \quad \\
  \textbf{Chengcheng Wan\textsuperscript{3} \quad
  Hongyu Zhang\textsuperscript{4} \quad
  Xiaodong Gu\textsuperscript{1}} \\
  \textsuperscript{1}Shanghai Jiao Tong University \quad
  \textsuperscript{2}Fudan University \quad \\
  \textsuperscript{3}East China Normal University \quad
  \textsuperscript{4}Chongqing University \\
  \texttt{\{zengwh\_cs, xiaodong.gu\}@sjtu.edu.cn}
}

\begin{document}
\maketitle
\begin{abstract}
Large Reasoning Models (LRMs) have demonstrated remarkable capabilities by scaling up the length of Chain-of-Thought (CoT). However, excessively long reasoning traces pose substantial challenges for training cost and inference latency.
While various CoT compression approaches have emerged to address this challenge, they face inherent trade-offs: token-level methods often disrupt syntactic and logical coherence, while step-level methods based on perplexity fail to reliably capture the logically critical reasoning steps because of the dilution of logical information.
In this paper, we propose \textbf{\approach} (\textbf{\underline{A}}nchor-guided, \textbf{\underline{S}}urpris\textbf{\underline{A}}l-based \textbf{\underline{P}}runing), a novel coarse-to-fine framework for CoT compression. 
\approach first performs anchor-guided pruning to preserve the core reasoning structure, which efficiently reduces the search space for subsequent processing.
Leveraging the insight that logical branching choices are concentrated at the onset of reasoning steps, it then enables logic-aware pruning by selecting logically essential reasoning steps based on a novel first-token surprisal metric. 
Finally, \approach distills the models to autonomously generate and leverage these concise CoTs at inference time, enabling efficient reasoning.
Experiments show that \approach achieves state-of-the-art accuracy across multiple benchmarks while substantially reducing training and inference costs.
\end{abstract}

\begin{figure}[t]
    \centering
    \includegraphics[width=\columnwidth,trim=0pt -5pt 0pt -5pt]{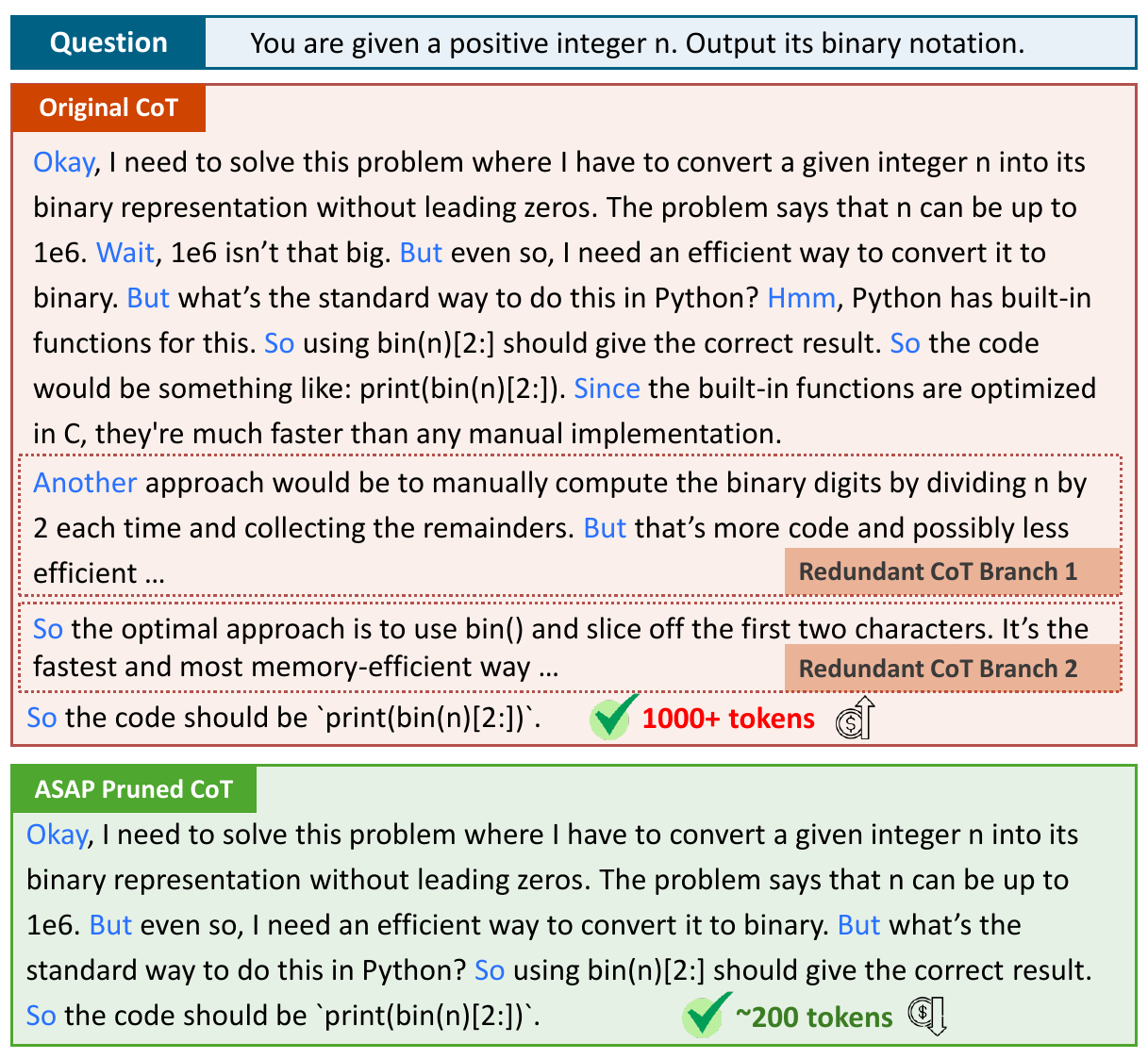}
    \caption{
    Illustration of CoT pruning by \emph{\approach{}}. 
    The \emph{Original CoT} generated by LRMs exhibits two types of redundancy: (1) \textbf{Structural Redundancy}, such as digressive branches (highlighted in red dashed boxes), which are removed by our Stage 1 Anchor-guided pruning; and (2) \textbf{Logical Redundancy} within valid paths. \emph{\approach} addresses the latter in Stage 2 by computing the surprisal of the first tokens of reasoning steps (marked in blue) to identify and retain only the critical cognitive pivots.
    }
    \label{fig:introduction}
\end{figure}

\section{Introduction}
\label{sec:introduction}
The emergence of Large Reasoning Models, including OpenAI's o1~\cite{jaech2024openai} and DeepSeek-R1~\cite{guo2025deepseek}, marks a paradigm shift in artificial intelligence. By scaling up Chain-of-Thought (CoT) reasoning~\cite{wei2022chain}, these models demonstrate emergent capabilities in complex domains such as mathematics~\cite{sun2025survey}, programming~\cite{shi2024code,yang2025elaboration,hu2025flowmaltrans}, and logical reasoning~\cite{liu2025logical,zhang2025gam,zhang2025mmcot}.
However, this performance comes at a prohibitive cost: reasoning traces often span thousands of tokens, introducing substantial latency and memory overhead. Crucially, these lengthy traces often contain substantial redundancy, such as over-explaining simple problems or superficially exploring multiple paths for complex ones~\cite{bi2024program, xie2025logic, wu2025more, qu2025survey}. 
For instance, the \emph{Original CoT} in Figure~\ref{fig:introduction} contains tangential branches (highlighted in red dashed boxes), such as exploring an alternative manual implementation that is subsequently rejected (``\textit{But that's more code...}''). Furthermore, the reasoning is punctuated by syntactic fillers that contribute little to the core logic.
This observation raises a fundamental question: \textit{Can we identify and retain only the ``cognitive pivots'' of reasoning while discarding the redundancy?}

A growing body of research has emerged on CoT compression for efficient reasoning~\cite{qu2025survey}. 
Token-level methods like TokenSkip~\cite{xia2025tokenskip} adapt general-purpose context compressors such as LLMLingua-2~\cite{pan2024llmlingua} to prune non-informative tokens. However, indiscriminate token removal risks disrupting the syntactic integrity of the reasoning chain. To address this, step-level pruning methods like SPIRIT~\cite{cui2025stepwise} trim entire reasoning steps, thereby preserving structural coherence. 
Nevertheless, these approaches face a fundamental challenge: accurately estimating the logical importance of each step. They typically rely on fixed metrics like perplexity (PPL), which measures the overall predictability of a sentence. This holistic measure often dilutes the signal of critical logical leaps with the noise of syntactically predictable but logically trivial content.

In this work, we ground CoT compression from an information-theoretic perspective. Through an empirical analysis of 10 million reasoning tokens (detailed in Section~\ref{sec:empirical_analysis}), we find that the logical progression within a CoT sequence is not uniformly distributed; instead, its information density is highly concentrated at the beginning of each reasoning step—specifically within the first few tokens (blue-highlighted in Figure~\ref{fig:introduction}).
These tokens, such as ``\textit{But}'' (self-correction) or ``\textit{So}'' (deduction, not continuation), serve as high-entropy \textbf{cognitive pivots}. By leveraging the surprisal of these initial tokens, we can distinguish between critical logical transitions and predictable elaborations.

Guided by this insight, we propose \textbf{\approach} (\textbf{\underline{A}}nchor-guided, \textbf{\underline{S}}urpris\textbf{\underline{A}}l-based \textbf{\underline{P}}runing), a coarse-to-fine framework designed to preserve these high-information steps.
\approach in a two-stage cascade that directly addresses the two types of redundancies identified in our study (illustrated in Figure~\ref{fig:introduction}):
First, it employs \textbf{Anchor-guided Pruning} to remove structural redundancies. By generating a concise step-by-step reasoning trace as a logical backbone, it identifies and prunes the irrelevant branches (e.g., the red boxes in Figure~\ref{fig:introduction}).
Second, it performs \textbf{Surprisal-based Refining} to eliminate logic-sparse steps. Leveraging our First-Token Surprisal metric, this stage iteratively filters out steps acting as mere fillers while retaining the high-surprisal cognitive pivots.
Finally, we distill these compact, logic-dense CoTs into a target model, enabling it to generate efficient reasoning chains.

We validate our approach through extensive experiments on the DeepSeek-R1-Distill-Qwen-7B and DeepSeek-R1-Distill-Llama-8B~\cite{guo2025deepseek} across diverse domains. The results demonstrate that \approach establishes a superior Pareto frontier between performance and efficiency. 
Notably, on the challenging LiveCodeBench v4\_v5 benchmark, \model achieves \textbf{36.19\%} Pass@1 while reducing token generation by \textbf{23.5\%} and inference latency by \textbf{43.5\%} compared to the strongest baseline.

Our main contributions are summarized as follows:
\begin{itemize}
    \itemsep0em
    \item We present an empirical analysis of the information concentration of CoTs, uncovering that the surprisal of the starting token for each CoT step is a more robust indicator of logical importance than perplexity.
    \item We propose \textbf{\approach}, a novel CoT compression framework that combines structural alignment with information-guided refinement.
    \item Extensive experiments on multiple benchmarks demonstrate that models fine-tuned on CoTs pruned by \approach achieve state-of-the-art accuracy while substantially reducing computational costs.
\end{itemize}

\begin{figure*}[t]
    \centering
    \begin{subfigure}{0.32\textwidth}
        \centering
        \includegraphics[width=\linewidth, height=3cm]{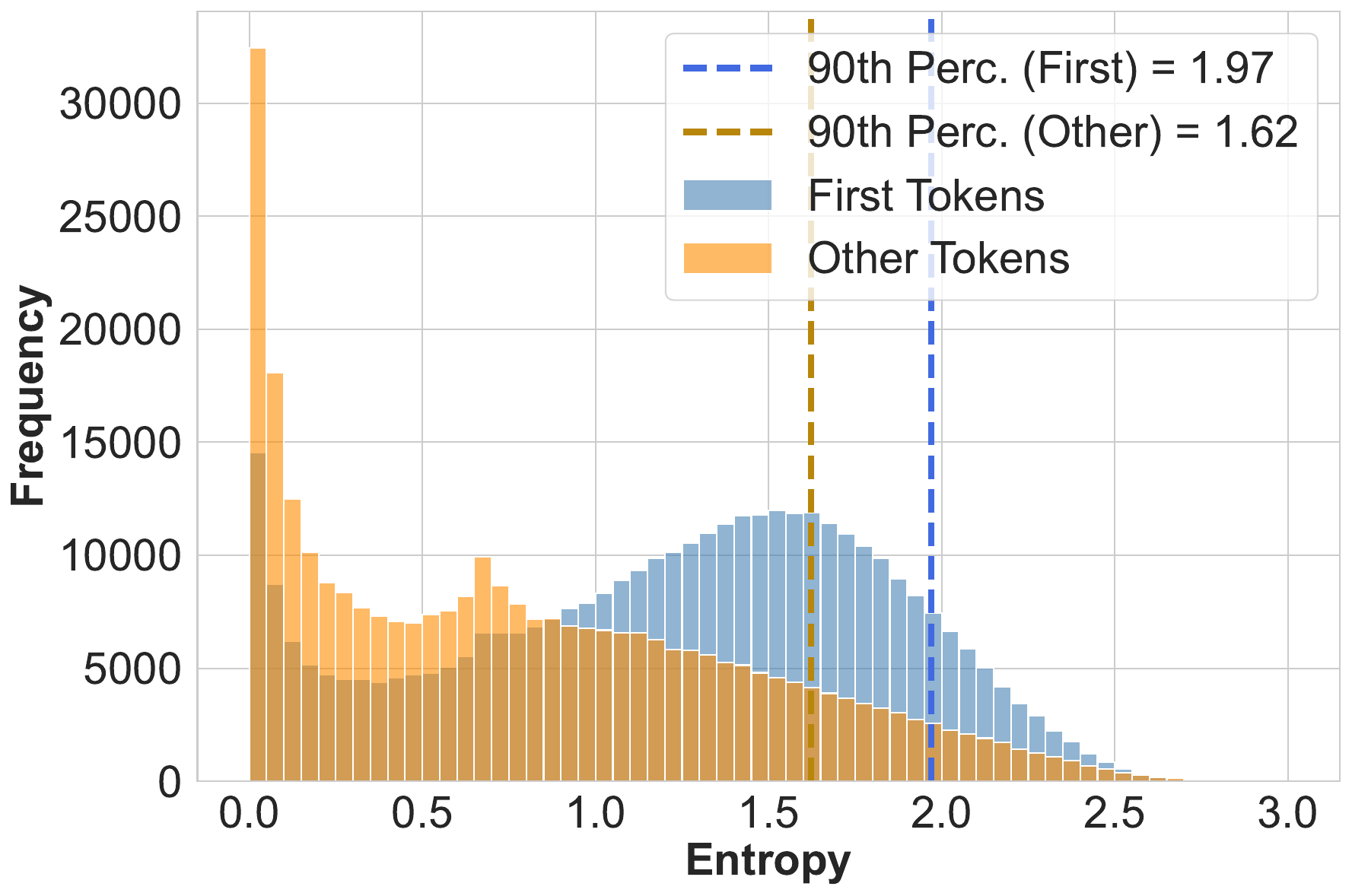}
        \caption{Entropy Distribution}
        \label{fig:entropy_dist}
    \end{subfigure}
    \hfill
    \begin{subfigure}{0.32\textwidth}
        \centering
        \includegraphics[width=\linewidth, height=3cm]{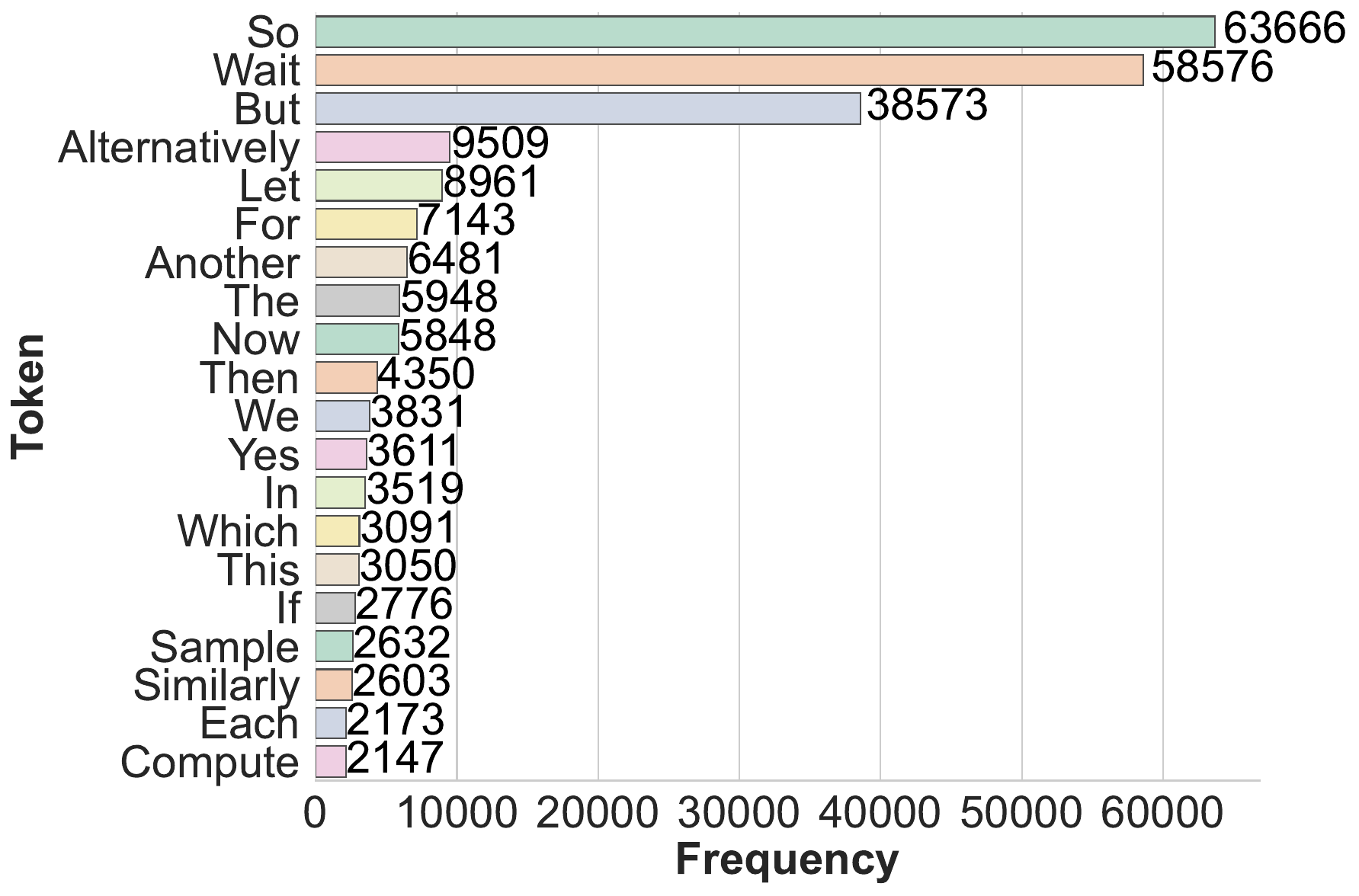}
        \caption{High-Frequency First Tokens}
        \label{fig:freq_cloud}
    \end{subfigure}
    \hfill
    \begin{subfigure}{0.32\textwidth}
        \centering
        \includegraphics[width=\linewidth, height=3cm]{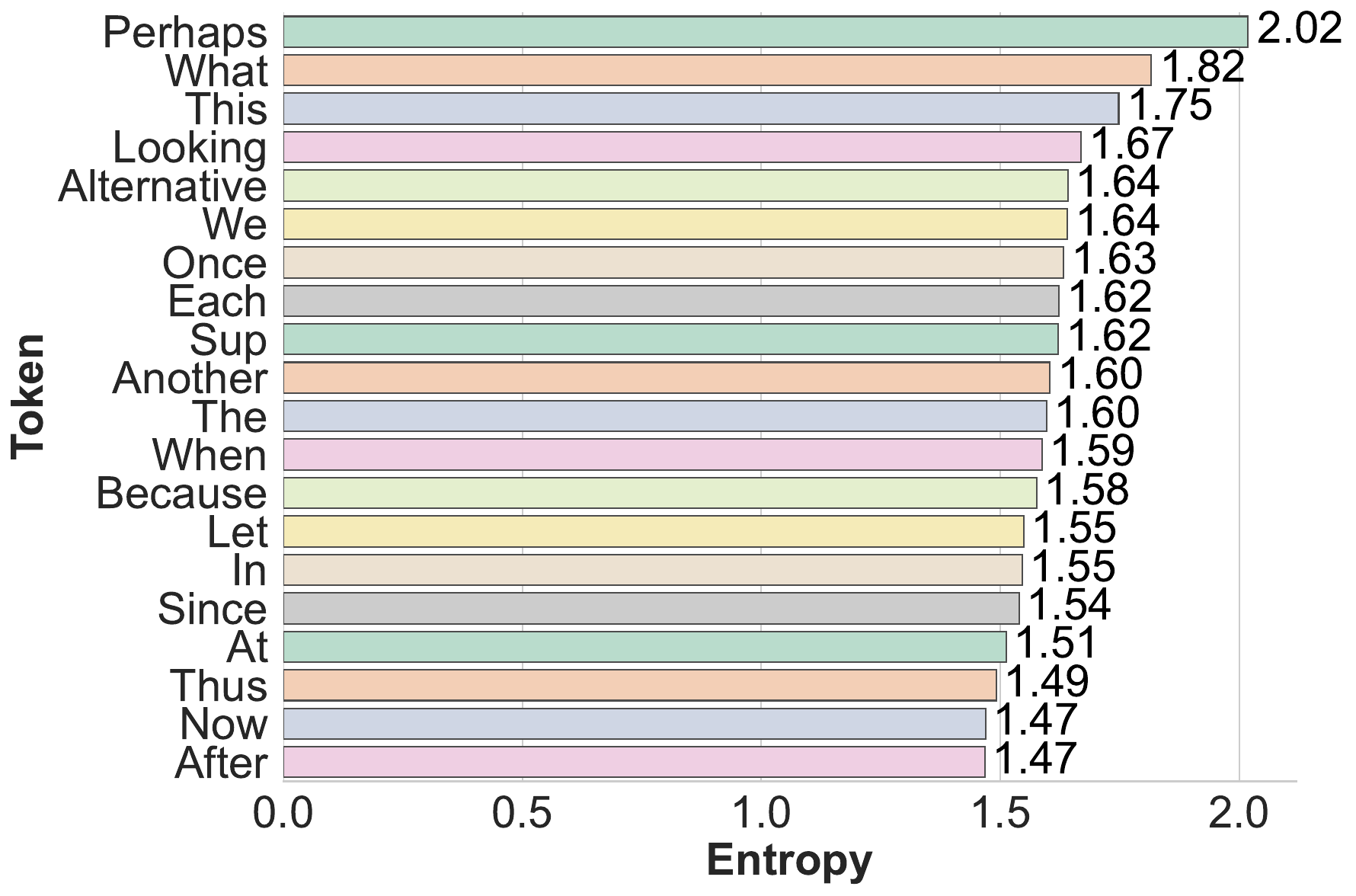}
        \caption{High-Entropy Tokens}
        \label{fig:surprisal_cloud}
    \end{subfigure}
    \caption{
    Empirical analysis of 10M tokens from DeepSeek-R1-Distill-Qwen-32B. 
    (a) The entropy distribution reveals a clear information concentration: first tokens (blue) exhibit significantly higher uncertainty (entropy) compared to body tokens (orange), which are highly deterministic. 
    (b) The most frequent first tokens are a mixture of logical operators (e.g., \textit{Wait}) and ubiquitous syntactic connectors (e.g., \textit{So}). 
    (c) High-entropy states filter out predictable fillers like \textit{So} or \textit{Then}, while exclusively highlighting cognitive pivots such as \textit{Perhaps}, \textit{What}, and \textit{Alternative}.
    }
    \label{fig:empirical_analysis}
\end{figure*}

\section{Empirical Analysis}
\label{sec:empirical_analysis}
To investigate the intrinsic distribution of logical information in CoTs, we conducted a large-scale analysis on 10 million tokens generated by DeepSeek-R1-Distill-Qwen-32B~\cite{guo2025deepseek} across diverse reasoning benchmarks (AIME and LiveCodeBench~\cite{jain2024livecodebench}).
\paragraph{Information Concentration in CoTs.}
We analyze the entropy distribution of the \textit{first token} of each reasoning step compared to all subsequent tokens. Entropy, in this context, quantifies the model's uncertainty regarding the next state transition~\cite{shannon1948mathematical, malinin2020uncertainty, kuhn2023semantic, wang2025beyond, cheng2025reasoning}. As illustrated in Figure~\ref{fig:empirical_analysis}(a), a distinct concentration is observed. Starting tokens (blue) exhibit a dispersed distribution with a significantly higher 90th percentile. In contrast, other tokens (orange) are heavily concentrated near zero entropy, indicating that once a reasoning step is initiated, its subsequent elaboration is largely deterministic and syntactically driven. This empirical evidence confirms that the logical branching points, where the model actively deliberates on the reasoning path, are structurally concentrated at the beginning of each step.

\paragraph{Identifying Cognitive Pivots with Entropy.}
Having identified the informative start tokens, we perform a more in-depth analysis of real logical pivots among the start tokens. We aim to distinguish between superficial syntactic connectors and real logical pivots. 
Figure~\ref{fig:empirical_analysis}(b) presents high-frequency start tokens that appear in CoTs, which mix connectors (``\textit{So}'', ``\textit{Let}'') with reasoning markers (``\textit{Wait}'', ``\textit{But}''). 
However, when focusing on tokens generated in high-entropy states (Figure~\ref{fig:empirical_analysis}(c)), a qualitative shift emerges. Predictable connectors like ``\textit{So}'' and ``\textit{Then}'' are effectively suppressed due to their low uncertainty. Instead, the distribution is dominated by terms representing cognitive pivots and state transitions, such as: 1) Exploration: ``\textit{Alternative}'', ``\textit{Another}'' (proposing hypotheses or new paths). 2) Causality: ``\textit{Because}'', ``\textit{Since}'' (providing formal justification). 3) Self-Correction: ``\textit{Perhaps}'', ``\textit{What}'' (indicating error detection or logic reversal).

This analysis demonstrates that high entropy is a robust indicator of logical salience. Since the actual next token is known in the given training sequence, we operationalize this insight by using \textit{First-Token Surprisal} as a proxy to identify and preserve these critical reasoning hops in our proposed framework~\cite{fu2025deep}.

\begin{figure*}[t]
    \centering
    \includegraphics[width=\textwidth]{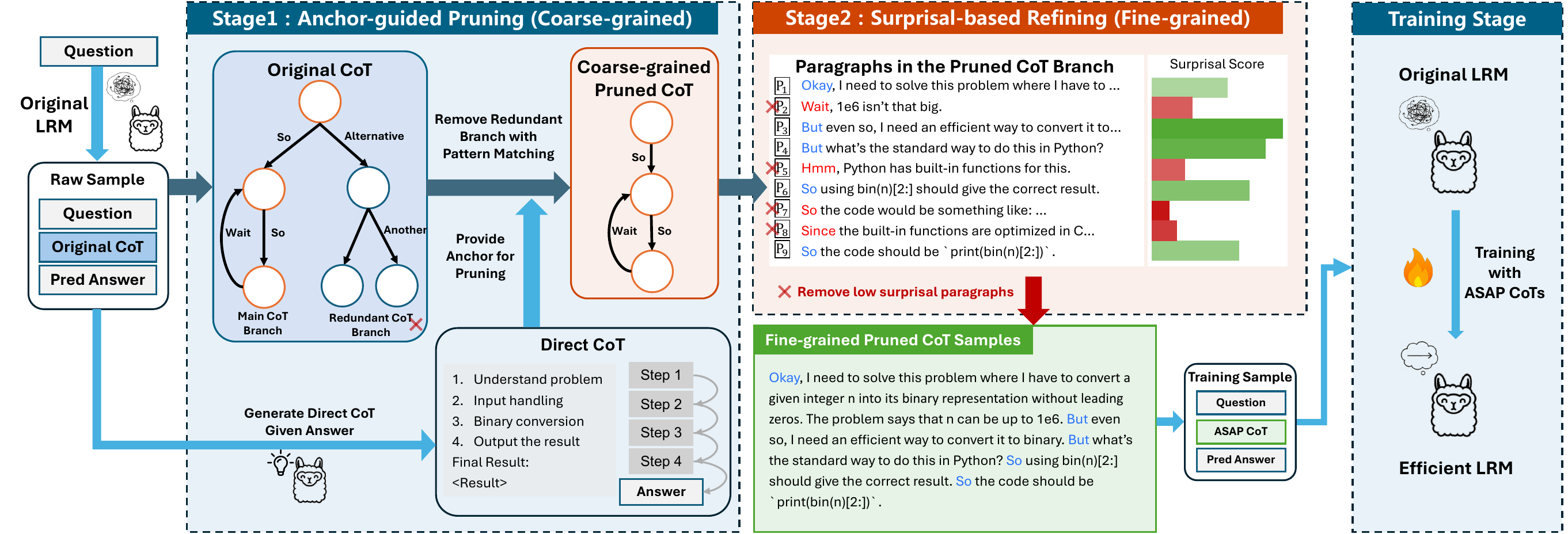}
    \caption{
    The overall framework of \textbf{\approach}. 
    The pipeline consists of three phases: 
    (1) \textbf{In Stage 1}, the LLM generates a ``Direct Thought" ($\mathcal{P}$) from the (Question, Answer) pair. \(\mathcal{P}\) acts as an anchor to prune the ``Original CoT" ($C$) into a ``Coarse-grained Pruned CoT" ($C_{coarse}$). 
    (2) \textbf{In Stage 2}, we compute the \textit{First-Token Surprisal} for each step in $C_{coarse}$. High-surprisal steps are retained, while low-surprisal fillers are pruned, yielding the final ``Fine-grained Pruned CoT" ($C'$).
    (3) \textbf{In Training Stage}, the data with \approach pruned CoTs is used to fine-tune the LRM for efficient inference.
    }
    \label{fig:framework}
\end{figure*}

\section{Methodology}
\label{sec:methodology}


\subsection{Overall Framework}

Formally, we consider a supervised reasoning task defined by a dataset $\mathcal{D} = \{(Q_i, C_i, A_i)\}_{i=1}^N$, where $Q_i$ is the query, $A_i$ is the predicted answer, and $C_i$ represents the original CoT generated by the LRMs. $C_i$ is a sequence of reasoning steps $C_i = \{s_1, s_2, \dots, s_L\}$.
Our goal is to compress each \(C_i\) into a concise pruned one \(C_i'\) such that \(|C_i'|\ll|C_i|\) while the LRMs maintains the quality of generated reasoning steps and answers when fine-tuned on the dataset $\mathcal{D'} = \{(Q_i, C_i', A_i)\}_{i=1}^N$.

We propose \textbf{\approach{}}, a coarse-to-fine framework tailored to the redundancy of ``Original CoT" ($C$), as illustrated in Figure~\ref{fig:framework}.
Stage 1 (Anchor-guided Pruning) reduces \textit{structural redundancy} (e.g., dead ends) by aligning the CoT with a generated logical backbone. The LLM generates a ``Direct Thought" ($\mathcal{P}$) from the $(Q, A)$ pairs. \(\mathcal{P}\) acts as an anchor to prune the $C$ into a ``Coarse-grained Pruned CoT" ($C_{coarse}$).
Stage 2 (Surprisal-based Refining) reduces \textit{logical redundancy} (e.g., syntactic fillers) by filtering non-informative steps. We approximate the information of each step in $C_{coarse}$ with their surprisal of start tokens and prune low-surprisal steps, yielding the final ``Fine-grained Pruned CoT" ($C'$).
Finally, all $\{(Q, C', A)\}$ are utilized to fine-tune the target model.

\subsection{Anchor-guided Pruning}
\label{sec:stage1}

Directly pruning raw CoTs is challenging due to the noise and unstructured digressions inherent in LLM reasoning~\cite{zhou2024can}. To address this, we first construct a high-level logical skeleton to narrow the pruning space.

\paragraph{Generate Direct Thoughts.}
We prompt the LLM to infer a concise reasoning path called ``Direct Thought'' ($\mathcal{P}$) based on the $(Q, A)$ pair (see Appendix~\ref{sec:prompts} for prompts). Unlike exploratory CoTs, $\mathcal{P}$ is generated as a structured, step-by-step explanation that outlines how to derive the answer from the question, exemplified in Appendix~\ref{sec:prompts}. This $\mathcal{P}$ acts as a reference anchor, outlining the least reasoning trajectory required to solve the problem.

\paragraph{Pruning with Pattern Matching.}
Guided by the anchor $\mathcal{P}$, we prompt the LLM to prune the original CoT $C$. Specifically, the LLM is instructed to: 1) remove unnecessary reasoning steps from \(C\); 2) retain all key supporting content that aligns with the logic of $\mathcal{P}$; and 3) crucially, preserve the original wording without introducing new information. 
The prompt used for pruning is shown in Appendix~\ref{sec:prompts}.
The goal is to extract the subsequence of $C$ that semantically aligns with $\mathcal{P}$ while discarding irrelevant branches (as shown in the ``Original CoT'' block of Figure~\ref{fig:framework}).

Crucially, to mitigate LLM hallucination during compression, we enforce an extractive constraint, which validates structural and semantic alignment with $C$.
Specifically, we design a pattern-matching algorithm that verifies whether each step in $C_{coarse}$ corresponds to a matching step in $C$ while preserving their original order. The matching is performed using Gestalt Pattern Matching~\cite{black2004ratcliff} as a text similarity metric. A pruning is considered valid only if all steps in $C_{coarse}$ achieve a similarity score above a predefined threshold $\tau$ when matched against sequential steps in $C$.
The full pattern-matching algorithm is detailed in Algorithm~\ref{alg:pattern_matching} (see Appendix~\ref{sec:algorithms}). 
We leverage high-temperature sampling, which provides the necessary diversity to efficiently re-prompt failed cases, ensuring that a valid $C_{\text{coarse}}$ can be eventually generated.

\subsection{Surprisal-based Refining}
\label{sec:stage2}
Following the coarse-grained pruning, the resulting $C_{coarse}$ may still contain verbose steps that contribute little to the logic. Grounded in our empirical finding that logical information is concentrated (Section~\ref{sec:empirical_analysis}), we perform a meticulous, logic-aware refinement in $C_{coarse}$ to identify more subtle redundancies within the core reasoning path. 

\paragraph{First-Token Surprisal as Logical Importance.}
\label{sec:surprisal}
We introduce \textit{First-Token Surprisal} as a novel metric to precisely quantify the logical importance of each step, enabling us to filter out the least informative ones and produce the final highly condensed CoT.
Let a reasoning step $s$ be a sequence of tokens $s = (x_1, x_2, \dots, x_T)$. The informational value of $s$ within the context of previous steps $\mathcal{C}_{pre}$ is typically estimated by its joint probability. However, our analysis reveals that the \textit{first token} $x_1$ serves as the ``cognitive pivot'' carrying the majority of the uncertainty.
Therefore, we define the \textit{First-Token Surprisal} $\mathcal{S}(s)$ as:
\begin{equation}
    \mathcal{S}(s \mid \mathcal{C}_{pre}) = - \log P_{\theta}(x_1 \mid \mathcal{C}_{pre})
    \label{eq:surprisal_def}
\end{equation}
where $P_{\theta}$ denotes the probability distribution of the LRM. A high $\mathcal{S}(s)$ indicates a high-information transition (e.g., initiating a new deduction or self-correction), whereas a low score suggests a deterministic continuation or syntactic filler.

\paragraph{Pruning using First-Token Surprisal.}
We formulate the fine-grained pruning as a constrained maximization problem. Our goal is to select a subset of steps $S' \subset C_{coarse}$ that maximizes the total logical information subject to a length budget $L_{max}$:
\begin{equation}
\begin{split}
    S^* = \operatorname*{arg\,max}_{S' \subseteq C_{coarse}} \sum_{s \in S'} \mathcal{S}(s) \\
    \text{s.t.} \quad \sum_{s \in S'} \text{len}(s) \le L_{max}
\end{split}
\end{equation}
This formulation explicitly prioritizes steps with high information density. To solve this efficiently, we employ a greedy iterative strategy. We calculate the surprisal score for all steps in $C_{coarse}$ and iteratively remove the step with the lowest $\mathcal{S}(s)$, while the relative order of steps in $S'$ is preserved. The detailed procedure is provided in Algorithm~\ref{alg:iterative_pruning} (see Appendix~\ref{sec:algorithms}).
This process yields the final fine-grained CoT $C'$, which retains the critical ``aha moments''~\cite{guo2025deepseek} while meeting efficiency constraints.

\subsection{Supervised Fine-tuning}
\label{sec:sft}

Following the pruning, we construct the final training dataset $\mathcal{D}' = \{(Q_i, \mathbf{C}_i', A_i)\}_{i=1}^N$. For each instance, we concatenate the pruned CoT ($\mathbf{C}_i'$) and the final answer ($A_i$) to form the complete target response $R_i$.
We then fine-tune the LRM to minimize the standard negative log-likelihood of the target response tokens, conditioned on the input question. Formally, the loss is defined as:
\begin{equation}
\mathcal{L} = - \sum_{i=1}^{N} \sum_{j=1}^{|R_i|} \log P_{\theta}(r_{i,j} | Q_i, r_{i,<j})
\label{eq:sft_loss}
\end{equation}
where $r_{i,j}$ is the $j$-th token of the target response $R_i$, and $\theta$ represents the parameters of the model being fine-tuned.
This supervised fine-tuning process effectively distills the knowledge from our pruning framework into the model. By training on these compact, logically salient examples, the model learns to internalize efficient reasoning patterns.

\section{Experiments}
\label{sec:experiments}

\input{tables/main_results}

\input{tables/main_results_math}


\subsection{Experimental Setup}
\label{sec:setup}

\paragraph{Models and Datasets.}
All experiments are conducted on the DeepSeek-R1-Distill-Qwen-7B and DeepSeek-R1-Distill-Llama-8B~\cite{guo2025deepseek}, with DeepSeek-R1-Distill-Qwen-7B as the default backbone across all settings.
For the code reasoning domains, we use the Python subset of the CodeForces-CoTs~\cite{openr1} dataset. For the math reasoning domain, we adopt the OpenR1-Math~\cite{openr1} dataset and randomly sample 10K instances to match the size of the code subset, ensuring a balanced comparison across domains. The datasets consist of high-quality Chain-of-Thought (CoT) samples generated by DeepSeek-R1, making it particularly suitable for training competitive reasoning tasks.
Detailed implementation settings (hyperparameters, hardware, etc.) are provided in Appendix~\ref{sec:implementation_details}.

\paragraph{Benchmarks.}

We evaluate our method on a suite of widely used benchmarks that cover both code generation and mathematical reasoning tasks. 
For code generation, we adopt HumanEval+~\cite{chen2021evaluating, liu2023your}, LiveCodeBench \texttt{v1\_v3}, LiveCodeBench \texttt{v4\_v5}~\cite{jain2024livecodebench}, and LeetCodeDataset~\cite{xia2025leetcodedataset}. 
For mathematical reasoning, we evaluate on GSM8K~\cite{cobbe2021training}, MATH500~\cite{hendrycksmath2021}, AIME24, and AIME25.

\paragraph{Baselines.}

We compare our method against a comprehensive set of baselines. 
\textbf{Zero-shot} refers to the original model without any task-specific fine-tuning. 
\textbf{Original} denotes the model fine-tuned on the uncompressed CoTs from the training data.
Among compression approaches, Selective Context~\cite{li2023compressing} prunes redundant lexical units based on self-information; 
\textbf{LLMLingua-2}~\cite{pan2024llmlingua} distills GPT-4’s token importance signals into a lightweight Transformer encoder trained as a token classifier; 
\textbf{TokenSkip}~\cite{xia2025tokenskip} learns to skip less informative tokens to achieve controllable compression; 
and \textbf{SPIRIT}~\cite{cui2025stepwise} identifies critical reasoning steps by measuring perplexity shifts.
Except for the zero-shot setting, all methods involve fine-tuning on CoTs processed according to their respective compression strategies.

\paragraph{Metrics.}
We evaluate both accuracy and inference efficiency of each approach across three metrics: \textbf{Pass@1 (Acc)}, which measures the percentage of problems correctly solved on the first attempt; \textbf{Tokens (Tok)}, which denotes the average number of tokens generated by the LRMs; and \textbf{Latency (Lat)}, which measures the average time (in seconds) required for the model generation.

\subsection{Main Results}
\label{sec:main_results}

Tables~\ref{tab:main_results} and~\ref{tab:math_results_math} present the results of various methods on all benchmarks. The results show that the model fine-tuned on CoTs pruned by \approach consistently achieves the best trade-off between accuracy and efficiency. It achieves the best accuracy while generating the fewest tokens, leading to the lowest generation latency.

We notice a clear distinction between token-level and step-level pruning strategies. Token-level baselines such as Selective Context, LLMLingua-2, and TokenSkip exhibit a significant performance degradation compared to the original CoTs. This is because the token removal disrupts the syntactic structure and semantic coherence of the original reasoning steps. Consequently, the fine-tuning data becomes fragmented and grammatically unnatural, making it difficult for the model to learn the intended logical flow of the CoT. Step-level methods, such as SPIRIT, perform significantly better than token-level pruning methods, due to the preservation of sentence-level integrity. While SPIRIT improves efficiency over the Original with comparable accuracy, our method achieves higher efficiency and accuracy at the same time. This improvement is particularly pronounced on the challenging LiveCodeBench \texttt{v4\_v5} benchmark: \model reduces the average number of generated tokens by \textbf{23.5\%} (from 7892 to 6035) and lowers generation latency by \textbf{43.5\%} (from 4.62s to 2.61s), while also achieving a \textbf{7.8\%} improvement in accuracy (Pass@1 increases from 33.58\% to 36.19\%).

\subsection{Ablation and Analysis}
\label{sec:ablation_study}

\paragraph{Effect of Different Components.}
\label{sec:components}

To validate the contribution of each component, we conduct an ablation study on three model variants. 1) \textit{w/o Anchor-guided Pruning}: which skips Stage 1 and applies only surprisal-based pruning to the original CoT. 2) \textit{w/o Surprisal-based Refining}: which omits the surprisal-based refinement stage; and 3) \textit{w/o Both Pruning}: equivalent to the original baseline, where the model is fine-tuned on the full, uncompressed CoT.
Table~\ref{tab:ablation_v4_v5} presents the results on LiveCodeBench v4\_v5, which is representative of the consistent trends observed across benchmarks. Additional results are included in Appendix~\ref{sec:ablation_study_details}.
The results show that both pruning stages are essential and mutually complementary for optimal accuracy and efficiency.
First, removing the anchor-guided pruning leads to a drop in both accuracy and efficiency. While the accuracy decrease is modest, the generation latency increases by a substantial \textbf{76.2\%} (from 2.61s to 4.60s), underscoring the importance of stage 1.
Second, removing the surprisal-based refining results in a significant degradation across all metrics. The accuracy drops by \textbf{12.4\%} (Pass@1 decreases from 36.19\% to 31.72\%) relative to the \approach, and efficiency improvements are largely lost. This highlights that our surprisal-based pruning mechanism is essential to select the most critical steps. 

\input{tables/ablation_v4_v5}


\paragraph{Generalization to Different Architectures.}
\label{sec:generalization}
To validate the generalizability of \approach, we replicate our main experiments on the DeepSeek-R1-Distill-Llama-8B. Following the same experimental protocol, we compare \model against three strong baselines: Zero-shot, Original, and SPIRIT. 
We observe consistent trends across all benchmarks. For brevity, we present representative results on two key benchmarks: LiveCodeBench v4\_v5 and AIME24 in Table~\ref{tab:part_results_llama}, while reporting the full results in the Appendix~\ref{sec:generalization_llama}.
The results in the Llama3.1 series are highly consistent with our findings in the Qwen2.5 series, confirming the generalization of the \approach. As shown in Table~\ref{tab:part_results_llama}, \approach achieves the highest accuracy on both benchmarks, and the efficiency improvements are even more pronounced.
On LiveCodeBench, for instance, \approach not only surpasses the accuracy of the Original baseline (32.84\% vs. 31.34\%) but also generates \textbf{49.1\%} fewer tokens and reduces latency by over \textbf{3x} (from 8.60s to 2.69s). 
This suggests that the \approach is particularly effective in identifying and distilling the core reasoning patterns, validating its robustness and broad applicability for improving reasoning efficiency across different model families.

\input{tables/part_results_llama}


\begin{figure}[t]
    \centering
    \includegraphics[width=0.9\columnwidth]{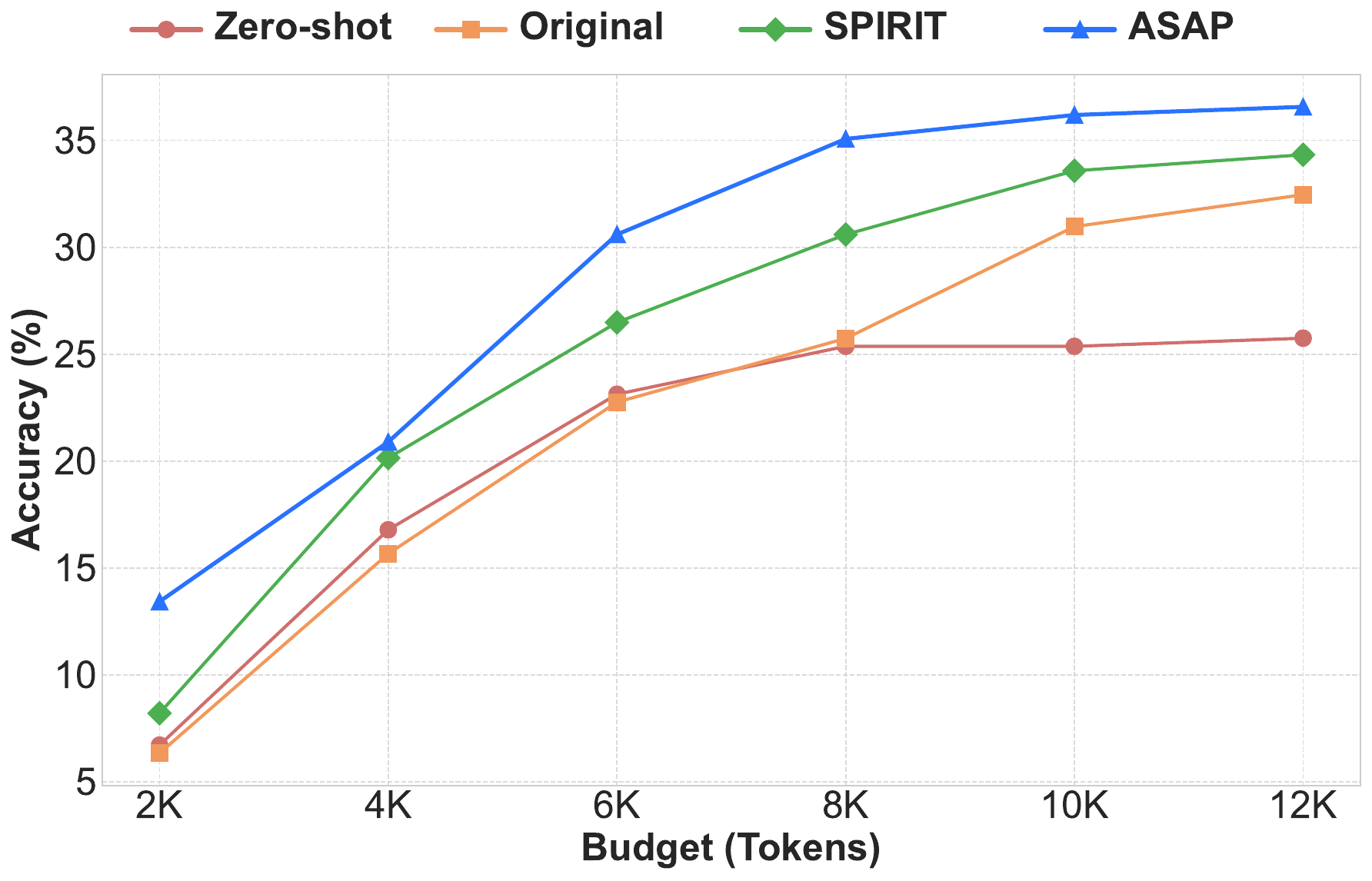}
    \caption{Performance of \approach on LiveCodeBench v4\_v5 under different token budgets.}
    \vspace{-10pt}
    \label{fig:budget_performance}
\end{figure}

\paragraph{Impact of Token Budget.}
\label{sec:token_budget}
To evaluate the scalability and resource sensitivity of our method, we analyze its behavior under varying inference-time token budgets (i.e., the maximum number of tokens to the model). We compare \approach against the three strong baselines-SPIRIT, Original, and Zero-shot—across all benchmarks, and observe consistent trends.
For clarity, we present results on LiveCodeBench v4\_v5 under six budget settings ranging from 2K to 12K tokens. Results for other benchmarks and additional statistics are provided in Appendix~\ref{sec:detailed_budgets}.
As shown in Figure~\ref{fig:budget_performance}, \approach consistently outperforms all baselines across all budget settings. 
In particular, \approach exhibits smooth performance scaling with respect to the token budget. 
We note that \approach achieves superior performance-efficiency trade-offs. For example, \approach with just an 8K token budget achieves higher accuracy than SPIRIT and Original at a much larger 12K budget.
These results further validate the practical utility of \approach in real-world scenarios.

\paragraph{Training Efficiency.}
\label{sec:training_efficiency}

To quantify the training efficiency gains, we present results of the CodeForces-CoTs dataset in Table~\ref{tab:training_results} and results on other datasets are provided in Appendix~\ref{sec:appendix_training_efficiency}.
The results highlight the training efficiency advantage of the \approach. By generating the most compact yet logically rich CoTs, our approach significantly reduces training overhead. Compared to the uncompressed baseline (Original), our method reduces the number of training tokens by \textbf{75.6\%} and shortens training time by \textbf{60.7\%}. These savings substantially exceed those achieved by all other baselines.
\approach enables a more resource-efficient training process, making it a practical and cost-effective solution for real-world deployment.

\input{tables/training_results}


\section{Related Work}
\label{sec:related_work}
\paragraph{Chain-of-Thought and Advanced Reasoning.}

Chain-of-Thought (CoT) prompting~\cite{wei2022chain} has evolved from heuristic prompting strategies~\cite{yao2023tree, lei2023boosting, ling2023deductive} to the training of specialized Large Reasoning Models (LRMs) like OpenAI's o1~\cite{jaech2024openai} and DeepSeek-R1~\cite{guo2025deepseek}. These models leverage reinforcement learning to scale test-time compute, generating lengthy reasoning traces to solve complex tasks~\cite{team2025kimi, yang2025qwen3, yu2025dapo, wang2025beyond,zhang2025cf,zhang2025tri}. 
Unlike prior works that enhance performance by scaling up CoT length, we focus on pruning redundancy to improve efficiency without compromising reasoning performance.

\paragraph{Context Compression for LLMs.}

To mitigate the computational cost of long contexts, various compression techniques have been proposed~\cite{zhang2025hybridtoken}. Approaches like Selective Context~\cite{li2023compressing}, LLMLingua series~\cite{jiang2023llmlingua, pan2024llmlingua}, and LongCodeZip~\cite{shi2025longcodezip} employ information-theoretic metrics or small external models to filter redundant tokens. 
However, these methods typically treat input as unstructured text. Applying them directly to CoT often disrupts the syntactic and logical coherence required for valid reasoning, a limitation that our pruning aims to overcome.

\paragraph{Efficient Reasoning via Fine-Tuning.}

Recent research has explored various efficiency mechanisms~\cite{qu2025survey}, ranging from compressing thoughts into continuous latent representations~\cite{hao2024training, cheng2024compressed, shen2025codi} to compressing CoTs~\cite{kang2025c3ot, xia2025tokenskip, cui2025stepwise}. 
Approaches like TokenSkip~\cite{xia2025tokenskip} and SPIRIT~\cite{cui2025stepwise} reduce length by filtering tokens or steps based on heuristics or perplexity shifts. 
However, these metrics often struggle to differentiate between syntactic fluency and logical necessity.
\approach differs by combining anchor-guided structural pruning with first-token surprisal, offering a more robust proxy for cognitive pivots.

\section{Conclusion}
\label{sec:conclusion}
In this paper, we address the inefficiency of Large Reasoning Models stemming from the structural and logical redundancies in Chain-of-Thought reasoning. 
Grounded in an information-theoretic perspective, our large-scale empirical analysis reveals a fundamental property of reasoning traces: Information Concentration, where the logical uncertainty is highly concentrated at the onset of reasoning steps.
Guided by this insight, we propose \approach. This coarse-to-fine framework first aligns the reasoning structure with a logical anchor and then refines it using a novel First-Token Surprisal metric.
Extensive experiments across multiple benchmarks demonstrate that \approach outperforms existing baselines, establishing a new state-of-the-art Pareto frontier between accuracy and efficiency.
Our work highlights the potential of using information-theoretic signals for efficient reasoning. Future work will explore applying \approach to online inference acceleration.

\section*{Limitations}
While \approach demonstrates significant improvements in reasoning efficiency, we acknowledge several limitations. 
First, our method relies on the availability of a capable LLM to generate high-quality ``Direct Thoughts'' in Stage 1. If the anchor contains logical errors or hallucinations, it may misguide the subsequent pruning, although our pattern-matching constraint mitigates this risk. 
Second, our experiments primarily focus on code generation and mathematical reasoning. While we believe the principle of information concentration applies broadly, the effectiveness of \approach on creative writing or commonsense reasoning tasks remains to be verified.




\bibliography{custom}


\appendix

\section{Algorithms}
\label{sec:algorithms}

\begin{algorithm}[H]
\small
\caption{Pattern Matching}
\label{alg:pattern_matching}
\begin{algorithmic}[1]
\Require Original CoT $C$, Coarse-grained Pruned CoT $C_{coarse}$, Threshold $\tau$
\Ensure \textit{True} if $C_{coarse}$ is valid, \textit{False} otherwise.

\Function{PatternMatch}{$C, C_{coarse}, \tau$}
    \State $S_{origin} \leftarrow \text{SplitStepsByBlankLine}(C)$
    \State $S_{coarse} \leftarrow \text{SplitStepsByBlankLine}(C_{coarse})$
    \State $origin\_idx \leftarrow 0$
    
    \For{each step $s_{coarse}$ in $S_{coarse}$}
        \State $found\_match \leftarrow \text{False}$
        \While{$origin\_idx < \text{Length}(S_{origin})$}
            \State $s_{origin} \leftarrow S_{origin}[origin\_idx]$
            \State $score \leftarrow \text{GestaltSimilarity}(s_{origin}, s_{coarse})$
            \If{$score \geq \tau$}
                \State $found\_match \leftarrow \text{True}$
                \State $origin\_idx \leftarrow origin\_idx + 1$
                \State \textbf{break}
            \EndIf
            \State $origin\_idx \leftarrow origin\_idx + 1$
        \EndWhile
        
        \If{\textbf{not} $found\_match$}
            \State \Return \textit{False}
        \EndIf
    \EndFor
    
    \State \Return \textit{True}
\EndFunction
\end{algorithmic}
\end{algorithm}

\begin{algorithm}[H]
\small
\caption{Iterative Pruning via First-Token Surprisal}
\label{alg:iterative_pruning}
\begin{algorithmic}[1]
\Require Coarse-grained Pruned CoT $C_{coarse}$, Max Tokens $L_{max}$, Model $M$, Tokenizer $T$
\Ensure Fine-grained Pruned CoT $C'$

\Function{FineGrainedPrune}{$C_{coarse}, L_{max}, M, T$}
    \If{$\text{Length}(T(C_{coarse})) \leq L_{max}$}
        \State \Return $C_{coarse}$
    \EndIf

    \State $S \leftarrow \text{SplitStepsByBlankLine}(C_{coarse})$
    \State $SurprisalScores \leftarrow \text{CalculateAll}(S, M, T)$
    
    \State $StepsToPrune \leftarrow \text{SortByScore}(S, \text{SurprisalScores})$
    
    \State $S_{current} \leftarrow S$
    \For{each step $s_{prune}$ in $StepsToPrune$}
        \State $S_{temp} \leftarrow S_{current} \setminus \{s_{prune}\}$
        \State $C_{temp} \leftarrow \text{Join}(S_{temp})$
        
        \If{$\text{Length}(T(C_{temp})) \leq L_{max}$}
            \State $S_{current} \leftarrow S_{temp}$
            \State \textbf{break}
        \EndIf
        \State $S_{current} \leftarrow S_{temp}$
    \EndFor
    
    \State $C' \leftarrow \text{Join}(S_{current})$
    \State \Return $C'$
\EndFunction

\end{algorithmic}
\end{algorithm}

\section{Prompt Templates and Generated Examples}
\label{sec:prompts}

We prompt the LLM to produce a direct thought: a concise, structured, step-by-step explanation that outlines how to derive the answer from the given question. A direct thought always consists of a small number of clear logical steps that directly contribute to the solution, and terminates with a single, explicit final answer. 
We use the prompt below to elicit such direct thoughts (the model is instructed to return only a detailed step-by-step solution containing only ``Step-by-Step Solution" and ``Final Answer").
\begin{tcolorbox}[
    colback=gray!10,  
    colframe=gray!60, 
    boxrule=0.5pt,    
    arc=2mm,          
    fonttitle=\bfseries, 
    coltitle=black,
    title={Prompt for Generating the Direct CoT \(\mathcal{P}\)} 
]

\small

Given a question, please tell me how to get this answer step by step. \\
Question: \\
\textasciigrave\textasciigrave\textasciigrave \\
\textit{\{question\}} \\
\textasciigrave\textasciigrave\textasciigrave \\
Answer: \\
\textasciigrave\textasciigrave\textasciigrave python \\
\textit{\{answer\}} \\
\textasciigrave\textasciigrave\textasciigrave \\
Only return a detailed step-by-step solution (containing only ``Step-by-Step Solution" and ``Final Answer"). The detailed step-by-step solution is: \\
\end{tcolorbox}

The following example illustrates a direct thought for a simple programming task: converting an integer into its binary representation. This example demonstrates the defining characteristics of direct CoTs, namely, concise step-by-step reasoning that directly derives the solution, without redundant exploration or verbose commentary.
\begin{tcolorbox}[
    colback=gray!10,
    colframe=gray!60,
    boxrule=0.5pt,
    arc=2mm,
    fonttitle=\bfseries,
    coltitle=black,
    title={Example of a Direct CoT}
]

\small

\textbf{Step-by-Step Solution} \\
1. Understand the Problem: The task is to convert a given positive integer \texttt{n} into its binary representation without leading zeros. The constraints are lenient (\(n \le 10^6\)), so even a straightforward approach is efficient.  \\
2. Input Handling: Read the integer \texttt{n} from standard input. Since the input consists of a single integer, we can directly use \texttt{input()} and convert it to an integer.  \\
3. Binary Conversion: Python’s built-in \texttt{bin()} function converts an integer to its binary string representation (prefixed with \texttt{0b}). To remove the prefix, slice the string from the third character onward (\texttt{[2:]}).  \\
4. Output the Result: Print the resulting binary string. This ensures no leading zeros are present.  \\
5. Efficiency: The approach is optimal because conversion and slicing both run in \(O(\log n)\) time, and memory usage is minimal. \\
\textbf{Final Code:} \\
\textasciigrave\textasciigrave\textasciigrave python \\
n = int(input()) \\
print(bin(n)[2:]) \\
\textasciigrave\textasciigrave\textasciigrave \\

\end{tcolorbox}

Given the direct thought as a reference, this stage instructs the model to prune redundant reasoning from the original chain of thought while preserving the essential logical structure. 
Specifically, the model is required to 1) remove unnecessary reasoning steps from the original CoT, 2) retain all key supporting content that aligns with the logic of direct CoT, and 3) strictly preserve the original wording and sentence order without introducing new information. 
This ensures that the compressed reasoning remains faithful to the original thought process while aligning with the concise, goal-oriented structure of the direct CoT. 
The following detailed prompt is used to elicit such coarse-grained pruning behavior.

\begin{tcolorbox}[
    colback=gray!10,  
    colframe=gray!60, 
    boxrule=0.5pt,    
    arc=2mm,          
    fonttitle=\bfseries, 
    coltitle=black,
    title={Prompt for Coarse-grained Pruning} 
]

\small

Compress the given thinking by referring to the provided solution. The goal is to remove irrelevant reasoning paths while retaining all content along the core reasoning path. Compression must be based on thinking, ensuring that the original wording and structure are preserved as much as possible. Follow these strict rules: \\
1. Use thinking as the foundation: Do not rewrite or replace its content with solution——only use solution to determine which parts are relevant. \\
2. Remove unnecessary reasoning: Aggressively remove alternative paths that are not part of the core reasoning path. \\
3. Retain key supporting content: Keep examples, reflections, and tests that help illustrate, verify, or analyze the core reasoning path. \\
4. Preserve original words: Do not paraphrase, reorder, or change any words. \\
5. Do not add new words: Do not introduce new concepts, symbols, or abbreviations. \\
If you understand, compress the following thinking based on the given solution. \\
Solution: \\
\textasciigrave\textasciigrave\textasciigrave \\
\textit{\{solution\}} \\
\textasciigrave\textasciigrave\textasciigrave \\
Thinking: \\
\textasciigrave\textasciigrave\textasciigrave \\
\textit{\{think\}} \\
\textasciigrave\textasciigrave\textasciigrave \\
The compressed thinking is: \\
\end{tcolorbox}

\section{Implementation Details}
\label{sec:implementation_details}
\paragraph{Software and Hardware.}
For fine-tuning, we utilized the unsloth library\footnote{https://pypi.org/project/unsloth/2025.5.6/} for its memory-efficient optimizations. For inference, we employed the vLLM engine\footnote{https://pypi.org/project/vllm/0.8.4/} to maximize throughput and efficiency. All experiments were conducted on NVIDIA H20 GPUs and Intel Xeon Platinum 8480+ CPUs.

\paragraph{Fine-tuning Configuration.}
We performed full-parameter fine-tuning for all models in our experiments. Key hyperparameters included precision set to bf16, \texttt{num\_train\_epochs} set to 10, and a \texttt{max\_seq\_length} of 16384. We used a \texttt{per\_device\_train\_batch\_size} of 1 with \texttt{gradient\_accumulation\_steps} set to 16, resulting in an effective batch size of 16. For the optimizer, we used AdamW with a \texttt{cosine\_with\_min\_lr} learning rate scheduler. The \texttt{warmup\_ratio} was set to 0.03, and the scheduler's \texttt{min\_lr\_rate} was 0.1 of the peak learning rate. To stabilize training, we applied gradient clipping with a \texttt{max\_grad\_norm} of 0.2. Based on preliminary experiments, we set the peak learning rate to $4 \times 10^{-5}$ for the DeepSeek-R1-Distill-Qwen-7B and $2 \times 10^{-5}$ for the DeepSeek-R1-Distill-Llama-8B. Due to the high computational cost of full-parameter fine-tuning, the model is fine-tuned by a single run with a fixed random seed 42.

\paragraph{Inference and Evaluation Protocol.}
All inference benchmarks were run using the vLLM engine with \texttt{dtype} set to bfloat16 and \texttt{gpu\_memory\_utilization} set to 0.9. To ensure deterministic and reproducible outputs, we set the sampling \texttt{temperature} to 0.0 and set \texttt{enable\_prefix\_caching} to False.
The default token budget for generation is adjusted based on the task difficulty. Specifically, it is 2K for GSM8K, 4K for MATH500, 6K for HumanEval+, and 10K for AIME24, AIME25, LiveCodeBench, and LeetCodeDataset. Results with other token budget settings are shown in Appendix~\ref{sec:detailed_budgets}.

\paragraph{Baseline Details.}
 Following established practices, we used a consistent scoring model; as our primary model is DeepSeek-R1-Distill checkpoints, we employed DeepSeek-R1-Distill-Qwen-7B for all model-scoring tasks. To ensure a fair comparison, we standardize the input format across all methods by preserving the original question and final answer, and applying compression only to the CoT reasoning steps. To balance compression ratio and content retention, we set the target compression ratio to 0.5 for all baseline methods, except for TokenSkip, where we follow its original design that allows a controllable compression ratio between 0.5 and 1.0.
Additionally, since the original SPIRIT method is computationally expensive when applied to extremely long CoTs, we adopt a modified version to ensure fair comparison: specifically, we compute perplexity once per reasoning step and iteratively remove steps until the target ratio is met. This variant retains the core idea of SPIRIT while improving scalability in our evaluation setting.

\paragraph{Hyperparameters for Our Method.}
Our method involves several stages. For the LLM-guided Coarse-grained Pruning stage, we employed DeepSeek-V3 for economic reasons. When generating the direct thought \(\mathcal{P}\), we used a deterministic setting (\texttt{temperature=0.0},  \texttt{top\_p=1.0}), while for making the final pruning result, we increased exploration (\texttt{temperature=1.0}, \texttt{top\_p=1.0}). For Pattern Matching, the similarity threshold \texttt{$\tau$} was set to 0.6. Finally, during Surprisal-based Fine-grained Pruning, the maximum token budget was set to 4096 to ensure a deep level of compression.

\section{Effect of Different Components.}
\label{sec:ablation_study_details}
To validate the contribution and necessity of each component in our two-stage pruning framework, we conduct a detailed ablation study. Specifically, we evaluate the following three variants: \textit{\model w/o Coarse-grained Pruning}, \textit{\model w/o Fine-grained Pruning}, and \textit{\model w/o Any Pruning}.
We present results on the HumanEval+, LiveCodeBench v1\_v3, and LeetCodeDatsets benchmarks in Table~\ref{tab:ablation_he+}, Table~\ref{tab:ablation_v1_v3}, and Table~\ref{tab:ablation_lcd}.


\input{tables/ablation_he+}

\input{tables/ablation_v1_v3}

\input{tables/ablation_lcd}

\section{Generalization to Different Architectures}
\label{sec:generalization_llama}
To evaluate the generalizability of \approach{}, we replicate our main experiments on the DeepSeek-R1-Distill-Llama-8B. Following the same experimental protocol, we compare \model against three baselines: Zero-shot, Original, and SPIRIT. The results of the code generation task on the HumanEval+, LiveCodeBench v1\_v3, LiveCodeBench v4\_v5, and LeetCodeDataset benchmarks are shown in Table~\ref{tab:main_results_llama}. The results of the mathematical reasoning task on the GSM8K, MATH500, AIME24, and AIME25 benchmarks are shown in Table~\ref{tab:main_results_llama_math}.

\input{tables/main_results_llama}

\input{tables/main_results_llama_math}

\section{Performance under Different Token Budgets}
\label{sec:detailed_budgets}
To evaluate the performance scalability and resource sensitivity of our method, we analyze its behavior under varying inference-time token budgets (i.e., the maximum number of tokens the model is allowed to generate). We compare \model with three strong baselines—SPIRIT, Original, and Zero-shot—on HumanEval+, LiveCodeBench v1\_v3, LiveCodeBench v4\_v5, LeetcodeDataset, GSM8K, MATH500, AIME24, and AIME25. For simpler benchmarks (including HumanEval+, GSM8K, and MATH500), we evaluate the performance under four budget settings, ranging from 1K to 6K tokens. For more complex benchmarks (including LiveCodeBench v1\_v3, LiveCodeBench v4\_v5, LeetcodeDataset, AIME24, and AIME25), we evaluate the performance under six budget settings, ranging from 2K to 12K tokens. Results are shown in Table~\ref{tab:budget_he+}, Table~\ref{tab:budget_lcbv1_v3}, Table~\ref{tab:budget_lcbv4_v5}, Table~\ref{tab:budget_lcd}, Table~\ref{tab:budget_gsm8k}, Table~\ref{tab:budget_math500}, Table~\ref{tab:budget_aime24}, and Table~\ref{tab:budget_aime25}.

\input{tables/budget_he+}

\input{tables/budget_lcb_v1_v3}

\input{tables/budget_lcb_v4_v5}

\input{tables/budget_lcd}

\input{tables/budget_gsm8k}

\input{tables/budget_math500}

\input{tables/budget_aime24}

\input{tables/budget_aime25}

\section{Training Efficiency}
\label{sec:appendix_training_efficiency}
To quantify the training efficiency gains, we present results of the CodeForces-CoTs dataset in Table~\ref{tab:training_results} and results of the OpenR1-Math dataset in Table~\ref{tab:training_results_math}. We report two key metrics: the \textit{average number of tokens} per sample and the \textit{average training time} measured in seconds per step.

\input{tables/training_results_math}

\end{document}

%% file: tables/main_results.tex
\begin{table*}[tbh]
\centering
\begin{tabular}{l@{\hspace{5pt}}c@{\hspace{5pt}}c@{\hspace{5pt}}c@{\hspace{5pt}}c@{\hspace{5pt}}c@{\hspace{5pt}}c@{\hspace{5pt}}c@{\hspace{5pt}}c@{\hspace{5pt}}c@{\hspace{5pt}}c@{\hspace{5pt}}c@{\hspace{5pt}}c}

\toprule

\multirow{2}{*}{\textbf{Methods}} & \multicolumn{3}{c}{\textbf{HE+}} & \multicolumn{3}{c}{\textbf{LCBv1\_v3}} & \multicolumn{3}{c}{\textbf{LCBv4\_v5}} & \multicolumn{3}{c}{\textbf{LCD}} \\
\cmidrule(lr){2-4} \cmidrule(lr){5-7} \cmidrule(lr){8-10} \cmidrule(lr){11-13} 
& Acc $\uparrow$ & Tok $\downarrow$ & Lat $\downarrow$ & Acc $\uparrow$ & Tok $\downarrow$ & Lat $\downarrow$ & Acc $\uparrow$ & Tok $\downarrow$ & Lat $\downarrow$ & Acc $\uparrow$ & Tok $\downarrow$ & Lat $\downarrow$ \\

\midrule
Zero-shot & 68.29 & 3051 & 1.16 & 42.16 & 7088 & 3.59 & 25.37 & 8336 & 5.15 & 19.74 & 8680 & 4.95 \\
Original & \underline{75.61} & 2973 & 1.12 & \underline{52.12} & 6611 & 3.15 & 30.97 & 8289 & 4.83 & \underline{25.00} & 8485 & 4.72 \\
\midrule 
Selective Context & 54.88 & 2979 & 1.13 & 30.23 & 7025 & 3.75 & 16.79 & 8558 & 5.35 & 15.79 & 8461 & 4.90 \\
LLMLingua-2 & 68.29 & 3075 & 1.19 & 38.89 & 6953 & 3.60 & 22.76 & 8474 & 5.31 & 17.54 & 8513 & 4.81 \\
TokenSkip & 73.78 & 2823 & \underline{1.07} & 32.35 & 7095 & 3.85 & 20.15 & 8400 & 5.37 & 18.42 & 8503 & 4.87 \\
SPIRIT & \underline{75.61} & \underline{2764} & \underline{1.07} & 50.82 & \underline{6524} & \underline{3.09} & \underline{33.58} & \underline{7892} & \underline{4.62} 
& \underline{25.00} & \underline{8186} & \underline{4.45} \\
\midrule 
\model & \textbf{78.66} & \textbf{2464} & \textbf{0.98} & \textbf{54.74} & \textbf{5177} & \textbf{2.09} & \textbf{36.19} & \textbf{6035} & \textbf{2.61} & \textbf{27.63} & \textbf{7541} & \textbf{3.48} \\
\bottomrule
\end{tabular}
\caption{Experimental results of different methods on code generation benchmarks with DeepSeek-R1-Distill-Qwen-7B. We report accuracy (Acc), average number of generated tokens (Tok), and average generation latency (Lat) measured in seconds. The best results are highlighted in bold, and the second-best are underlined.}
\label{tab:main_results}
\end{table*}

%% file: tables/main_results_math.tex
\begin{table*}[tbh]
\centering
\begin{tabular}{l@{\hspace{5pt}}c@{\hspace{5pt}}c@{\hspace{5pt}}c@{\hspace{5pt}}c@{\hspace{5pt}}c@{\hspace{5pt}}c@{\hspace{5pt}}c@{\hspace{5pt}}c@{\hspace{5pt}}c@{\hspace{5pt}}c@{\hspace{5pt}}c@{\hspace{5pt}}c}
\toprule
\multirow{2}{*}{\textbf{Methods}} & \multicolumn{3}{c}{\textbf{GSM8K}} & \multicolumn{3}{c}{\textbf{MATH500}} & \multicolumn{3}{c}{\textbf{AIME24}} & \multicolumn{3}{c}{\textbf{AIME25}} \\
\cmidrule(lr){2-4} \cmidrule(lr){5-7} \cmidrule(lr){8-10} \cmidrule(lr){11-13}
& Acc $\uparrow$ & Tok $\downarrow$ & Lat $\downarrow$ & Acc $\uparrow$ & Tok $\downarrow$ & Lat $\downarrow$ & Acc $\uparrow$ & Tok $\downarrow$ & Lat $\downarrow$ & Acc $\uparrow$ & Tok $\downarrow$ & Lat $\downarrow$ \\
\midrule
Zero-shot & 83.55 & 1301 & 0.27 & 60.40 & 2629 & 0.70 & \underline{36.67} & 8352 & 6.76 & \underline{40.00} & 8145 & 6.67 \\
Original & 86.35 & 1250 & 0.26 & 63.80 & 2511 & 0.66 & \textbf{46.67} & 8034 & 6.43 & \textbf{43.33} & 8026 & 6.75 \\
\midrule 
Selective Context & 75.44 & \underline{1108} & 0.24 & 52.20 & 2507 & 0.66 & 16.67 & 9610 & 7.40 & 10.00 & 9329 & 7.40 \\
LLMLingua-2 & 79.98 & 1128 & 0.24 & 54.60 & 2802 & 0.76 & \underline{36.67} & 8369 & 6.60 & 23.33 & 8919 & 7.42 \\
TokenSkip & 85.37 & 1303 & 0.27 & \underline{65.60} & 2483 & 0.65 & \underline{36.67} & 8073 & 6.61 & 33.33 & 8465 & 7.44 \\
SPIRIT & \underline{88.55} & 1118 & \underline{0.23} & 64.20 & \underline{2144} & \underline{0.57} & \textbf{46.67} & \underline{7198} & \underline{5.78} & \textbf{43.33} & \underline{7817} & \underline{6.57} \\
\midrule 
\model & \textbf{90.75} & \textbf{753} & \textbf{0.16} & \textbf{70.80} & \textbf{1649} & \textbf{0.43} & \textbf{46.67} & \textbf{5552} & \textbf{5.04} & 36.67 & \textbf{5434} & \textbf{5.10} \\
\bottomrule
\end{tabular}
\caption{Experimental results of different methods on mathematical reasoning benchmarks with DeepSeek-R1-Distill-Qwen-7B. We report accuracy (Acc), average number of generated tokens (Tok), and average generation latency (Lat) measured in seconds. The best results are highlighted in bold, and the second-best are underlined.}
\vspace{-10pt}
\label{tab:math_results_math} 
\end{table*}

%% file: tables/ablation_v4_v5.tex
\begin{table}[t]
\centering
\begin{tabular}{l@{\hspace{2pt}}c@{\hspace{2pt}}c@{\hspace{2pt}}c}
    \toprule
    \textbf{Variants} & \textbf{Acc $\uparrow$} & \textbf{Tok $\downarrow$} & \textbf{Lat $\downarrow$} \\
    \midrule
    \textbf{\model} & \textbf{36.19} & \textbf{6035} & \textbf{2.61} \\
    w/o Anchor-guided Pruning & 35.07 & 7735 & 4.60 \\
    w/o Surprisal-based Refining & 31.72 & 8061 & 4.83 \\
    w/o Both Pruning & 30.97 & 8289 & 4.83 \\
    \bottomrule
\end{tabular}
\caption{Ablation study of different pruning strategies on LiveCodeBench v4\_v5. We report accuracy (Acc), average number of generated tokens (Tok), and average generation latency (Lat) measured in seconds.}
\label{tab:ablation_v4_v5}
\end{table}
\vspace{-10pt}

%% file: tables/part_results_llama.tex










\begin{table}[t]
\centering

\begin{tabular}
{l@{\hspace{2pt}}c@{\hspace{2pt}}c@{\hspace{2pt}}c@{\hspace{2pt}}c@{\hspace{2pt}}c@{\hspace{2pt}}c}
\toprule

\multirow{2}{*}{\textbf{Methods}} & \multicolumn{3}{c}{\textbf{LCB}} & \multicolumn{3}{c}{\textbf{AIME}} \\
\cmidrule(lr){2-4} \cmidrule(lr){5-7}

& Acc $\uparrow$ & Tok $\downarrow$ & Lat $\downarrow$ & Acc $\uparrow$ & Tok $\downarrow$ & Lat $\downarrow$ \\
\midrule

Zero-shot & 25.00 & 8508 & 8.90 & 33.33 & 8445 & 10.42 \\

Original & 31.34 & 8202 & 8.60 & \textbf{36.67} & 8550 & 10.04 \\

SPIRIT & 30.22 & 7913 & 8.45 & \textbf{36.67} & 8788 & 10.04 \\

\midrule

\model & \textbf{32.84} & \textbf{4175} & \textbf{2.69} & \textbf{36.67} & \textbf{5314} & \textbf{6.97} \\

\bottomrule
\end{tabular}
\caption{Experimental results of different methods with DeepSeek-R1-Distill-Llama-8B. We report accuracy (Acc), average number of generated tokens (Tok), and average generation latency (Lat) measured in seconds. The best results are highlighted in bold.}
\label{tab:part_results_llama}
\end{table}
\vspace{-10pt}

%% file: tables/training_results.tex
\begin{table}[t]
\centering
\begin{tabular}{l@{\hspace{2pt}}c@{\hspace{2pt}}c}
\toprule
\textbf{Methods} & \textbf{Tokens} & \textbf{Time} \\
\midrule
Original & 13023 & 80.11 \\
\midrule
Selective Context & 6722 (-48.4\%) & 63.41 (-20.9\%) \\
LLMLingua-2 & 6919 (-46.9\%) & 65.25 (-18.6\%) \\
TokenSkip & 9813 (-24.6\%) & 77.27 (-3.6\%) \\
SPIRIT & 6082 (-53.3\%) & 57.45 (-28.3\%) \\
\midrule
\model & \textbf{3178 (-75.6\%)} & \textbf{31.48 (-60.7\%)} \\
\bottomrule
\end{tabular}
\caption{Training efficiency comparison on CodeForces-CoTs dataset. We report the average number of tokens per sample and training time measured in seconds per step. Percentages indicate the reduction relative to the Original baseline.}
\vspace{-10pt}
\label{tab:training_results}
\end{table}

%% file: tables/ablation_he+.tex
\begin{table}[tbh]
\centering
    \begin{tabular}{l@{\hspace{2pt}}c@{\hspace{2pt}}c@{\hspace{2pt}}c}
        \toprule
        \textbf{Variants} & \textbf{Acc $\uparrow$} & \textbf{Tok $\downarrow$} & \textbf{Lat $\downarrow$} \\
        \midrule
        \textbf{\model} & \textbf{78.66} & \textbf{2464} & \textbf{0.98} \\
        w/o Coarse-grained Pruning & 78.05 & 2839 & 1.10 \\
        w/o Fine-grained Pruning & 67.07 & 2897 & 1.10 \\
        w/o Any Pruning & 75.61 & 2973 & 1.12 \\
        \bottomrule
    \end{tabular}
\caption{Ablation study of different pruning strategies for \model on HumanEval+. We report accuracy (Acc), average number of generated tokens (Tok), and average generation latency (Lat) measured in seconds.}
\label{tab:ablation_he+}
\end{table}

%% file: tables/ablation_v1_v3.tex
\begin{table}[tbh]
\centering
    \begin{tabular}{l@{\hspace{2pt}}c@{\hspace{2pt}}c@{\hspace{2pt}}c}
        \toprule
        \textbf{Variants} & \textbf{Acc $\uparrow$} & \textbf{Tok $\downarrow$} & \textbf{Lat $\downarrow$} \\
        \midrule
        \textbf{\model} & \textbf{54.74} & \textbf{5177} & \textbf{2.09} \\
        w/o Coarse-grained Pruning & 53.92 & 6107 & 2.77\\
        w/o Fine-grained Pruning & 51.14 & 6599 & 3.20 \\
        w/o Any Pruning & 52.12 & 6611 & 3.15 \\
        \bottomrule
    \end{tabular}
\caption{Ablation study of different pruning strategies for \model on LiveCodeBench v1\_v3. We report accuracy (Acc), average number of generated tokens (Tok), and average generation latency (Lat) measured in seconds.}
\label{tab:ablation_v1_v3}
\end{table}

%% file: tables/ablation_lcd.tex
\begin{table}[tbh]
\centering
    \begin{tabular}{l@{\hspace{2pt}}c@{\hspace{2pt}}c@{\hspace{2pt}}c}
        \toprule
        \textbf{Variants} & \textbf{Acc $\uparrow$} & \textbf{Tok $\downarrow$} & \textbf{Lat $\downarrow$} \\
        \midrule
        \textbf{\model} & \textbf{27.63} & \textbf{7541} & \textbf{3.48} \\
        w/o Coarse-grained Pruning & 24.12 & 7954 & 3.75 \\
        w/o Fine-grained Pruning & 25.44 & 8326 & 4.77 \\
        w/o Any Pruning & 25.00 & 8485 & 4.72 \\
        \bottomrule
    \end{tabular}
\caption{Ablation study of different pruning strategies for \model on LeetCodeDataset. We report accuracy (Acc), average number of generated tokens (Tok), and average generation latency (Lat) measured in seconds.}
\label{tab:ablation_lcd}
\end{table}

%% file: tables/main_results_llama.tex
\begin{table*}[ht]
\centering

\begin{tabular}
{l@{\hspace{5pt}}c@{\hspace{5pt}}c@{\hspace{5pt}}c@{\hspace{5pt}}c@{\hspace{5pt}}c@{\hspace{5pt}}c@{\hspace{5pt}}c@{\hspace{5pt}}c@{\hspace{5pt}}c@{\hspace{5pt}}c@{\hspace{5pt}}c@{\hspace{5pt}}c}

\toprule

\multirow{2}{*}{\textbf{Methods}} & \multicolumn{3}{c}{\textbf{HE+}} & \multicolumn{3}{c}{\textbf{LCBv1\_v3}} & \multicolumn{3}{c}{\textbf{LCBv4\_v5}} & \multicolumn{3}{c}{\textbf{LCD}} \\
\cmidrule(lr){2-4} \cmidrule(lr){5-7} \cmidrule(lr){8-10} \cmidrule(lr){11-13} 
& Acc $\uparrow$ & Tok $\downarrow$ & Lat $\downarrow$ & Acc $\uparrow$ & Tok $\downarrow$ & Lat $\downarrow$ & Acc $\uparrow$ & Tok $\downarrow$ & Lat $\downarrow$ & Acc $\uparrow$ & Tok $\downarrow$ & Lat $\downarrow$ \\

\midrule

Zero-shot & 64.02 & 3334 & 1.86 & 44.12 & 7162 & 6.92 & 25.00 & 8508 & 8.90 & 27.19 & 8358 & 8.65 \\

Original & 76.22 & 2978 & 1.63 & 52.61 & 6614 & 6.16 & 31.34 & 8202 & 8.60 & 26.32 & 8413 & 8.85 \\

SPIRIT & 72.56 & 3159 & 1.74 & 52.61 & 6280 & 5.84 & 30.22 & 7913 & 8.45 & 26.75 & 8449 & 8.73 \\

\midrule

\model & 76.83 & 2494 & 1.30 & 48.86 & 3605 & 2.18 & 32.84 & 4175 & 2.69 & 27.63 & 3792 & 2.42 \\

\bottomrule
\end{tabular}
\caption{Experimental results of different methods on code generation benchmarks with DeepSeek-R1-Distill-Llama-8B. We report accuracy (Acc), average number of generated tokens (Tok), and average generation latency (Lat) measured in seconds.}
\label{tab:main_results_llama}
\end{table*}

%% file: tables/main_results_llama_math.tex
\begin{table*}[ht]
\centering

\begin{tabular}
{l@{\hspace{5pt}}c@{\hspace{5pt}}c@{\hspace{5pt}}c@{\hspace{5pt}}c@{\hspace{5pt}}c@{\hspace{5pt}}c@{\hspace{5pt}}c@{\hspace{5pt}}c@{\hspace{5pt}}c@{\hspace{5pt}}c@{\hspace{5pt}}c@{\hspace{5pt}}c}

\toprule

\multirow{2}{*}{\textbf{Methods}} & \multicolumn{3}{c}{\textbf{GSM8K}} & \multicolumn{3}{c}{\textbf{MATH500}} & \multicolumn{3}{c}{\textbf{AIME24}} & \multicolumn{3}{c}{\textbf{AIME25}} \\
\cmidrule(lr){2-4} \cmidrule(lr){5-7} \cmidrule(lr){8-10} \cmidrule(lr){11-13} 
& Acc $\uparrow$ & Tok $\downarrow$ & Lat $\downarrow$ & Acc $\uparrow$ & Tok $\downarrow$ & Lat $\downarrow$ & Acc $\uparrow$ & Tok $\downarrow$ & Lat $\downarrow$ & Acc $\uparrow$ & Tok $\downarrow$ & Lat $\downarrow$ \\

\midrule

Zero-shot & 79.15 & 1262 & 0.36 & 57.20 & 2612 & 1.08 & 33.33 & 8445 & 10.42 & 26.67 & 8597 & 10.54 \\

Original & 84.91 & 1310 & 0.37 & 63.00 & 2534 & 1.01 & 36.67 & 8550 & 10.04 & 30.00 & 8268 & 10.05 \\

SPIRIT & 85.67 & 1256 & 0.35 & 62.60 & 2533 & 1.01 & 36.67 & 8788 & 10.04 & 36.67 & 8094 & 9.57 \\

\midrule 

\model & 87.34 & 768 & 0.20 & 66.00 & 1734 & 0.65 & 36.67 & 5314 & 6.97 & 33.33 & 5348 & 7.05 \\

\bottomrule
\end{tabular}
\caption{Experimental results of different methods on mathematical reasoning benchmarks with DeepSeek-R1-Distill-Llama-8B. We report accuracy (Acc), average number of generated tokens (Tok), and average generation latency (Lat) measured in seconds.}
\label{tab:main_results_llama_math}
\end{table*}

%% file: tables/budget_he+.tex
\begin{table*}[tbh]
\centering

    \centering
    \begin{tabular}{l@{\hspace{5pt}}c@{\hspace{5pt}}c@{\hspace{5pt}}c@{\hspace{5pt}}c@{\hspace{5pt}}c@{\hspace{5pt}}c@{\hspace{5pt}}c@{\hspace{5pt}}c@{\hspace{5pt}}c@{\hspace{5pt}}c@{\hspace{5pt}}c@{\hspace{5pt}}c}
      \toprule
      \multirow{2}{*}{\textbf{Budget}} & \multicolumn{3}{c}{\textbf{Zero-shot}} & \multicolumn{3}{c}{\textbf{Original}} & \multicolumn{3}{c}{\textbf{SPIRIT}} & \multicolumn{3}{c}{\textbf{\model}} \\
      \cmidrule(lr){2-4} \cmidrule(lr){5-7} \cmidrule(lr){8-10} \cmidrule(lr){11-13}
      & Acc $\uparrow$ & Tok $\downarrow$ & Lat $\downarrow$ & Acc $\uparrow$ & Tok $\downarrow$ & Lat $\downarrow$ & Acc $\uparrow$ & Tok $\downarrow$ & Lat $\downarrow$ & Acc $\uparrow$ & Tok $\downarrow$ & Lat $\downarrow$ \\
      \midrule
      \textbf{1K} & 9.76  & 1007 & 0.28 & 14.63 & 983  & 0.28 & 10.98 & 995  & 0.28 & 23.78 & 946  & 0.27 \\
      \textbf{2K} & 42.68 & 1813 & 0.53 & 43.29 & 1702 & 0.49 & 47.56 & 1690 & 0.49 & 54.88 & 1502 & 0.44 \\
      \textbf{4K} & 66.46 & 2561 & 0.85 & 65.85 & 2511 & 0.82 & 69.51 & 2401 & 0.80 & 71.34 & 2116 & 0.72 \\
      \textbf{6K} & 68.29 & 3051 & 1.16 & 75.61 & 2973 & 1.12 & 75.61 & 2764 & 1.07 & 78.66 & 2464 & 0.98 \\
      \bottomrule
    \end{tabular}%

\caption{Results of different methods under different budgets on HumanEval+. We report accuracy (Acc), average number of generated tokens (Tok), and average generation latency (Lat) measured in seconds.}
\label{tab:budget_he+}
\end{table*}

%% file: tables/budget_lcb_v1_v3.tex
\begin{table*}[tbh]
\centering

    \centering
 .  \begin{tabular}{l@{\hspace{5pt}}c@{\hspace{5pt}}c@{\hspace{5pt}}c@{\hspace{5pt}}c@{\hspace{5pt}}c@{\hspace{5pt}}c@{\hspace{5pt}}c@{\hspace{5pt}}c@{\hspace{5pt}}c@{\hspace{5pt}}c@{\hspace{5pt}}c@{\hspace{5pt}}c}
      \toprule
      \multirow{2}{*}{\textbf{Budget}} & \multicolumn{3}{c}{\textbf{Zero-shot}} & \multicolumn{3}{c}{\textbf{Original}} & \multicolumn{3}{c}{\textbf{SPIRIT}} & \multicolumn{3}{c}{\textbf{\model}} \\
      \cmidrule(lr){2-4} \cmidrule(lr){5-7} \cmidrule(lr){8-10} \cmidrule(lr){11-13}
      & Acc $\uparrow$ & Tok $\downarrow$ & Lat $\downarrow$ & Acc $\uparrow$ & Tok $\downarrow$ & Lat $\downarrow$ & Acc $\uparrow$ & Tok $\downarrow$ & Lat $\downarrow$ & Acc $\uparrow$ & Tok $\downarrow$ & Lat $\downarrow$ \\
      \midrule
      \textbf{2K}   & 16.50 & 1966 & 0.52 & 17.16 & 1920 & 0.51 & 18.95 & 1908 & 0.51 & 21.57 & 1833 & 0.49 \\
      \textbf{4K}   & 32.68 & 3499 & 1.06 & 30.72 & 3432 & 1.05 & 34.80 & 3370 & 1.03 & 34.97 & 3244 & 1.00 \\
      \textbf{6K}   & 39.05 & 4806 & 1.70 & 42.65 & 4673 & 1.67 & 43.14 & 4605 & 1.64 & 46.24 & 4358 & 1.54 \\
      \textbf{8K}   & 44.28 & 5903 & 2.46 & 47.71 & 5723 & 2.43 & 51.80 & 5515 & 2.27 & 52.61 & 4919 & 1.90 \\
      \textbf{10K}  & 42.16 & 7088 & 3.59 & 52.12 & 6611 & 3.15 & 50.82 & 6524 & 3.09 & 54.74 & 5177 & 2.09 \\
      \textbf{12K}  & 43.95 & 7988 & 5.10 & 54.41 & 7473 & 4.22 & 51.63 & 7362 & 4.09 & 55.56 & 5322 & 2.27 \\
      \bottomrule
    \end{tabular}%

\caption{Results of different methods under different budgets on LiveCodeBench v1\_v3. We report accuracy (Acc), average number of generated tokens (Tok), and average generation latency (Lat) measured in seconds.}
\label{tab:budget_lcbv1_v3}
\end{table*}

%% file: tables/budget_lcb_v4_v5.tex
\begin{table*}[tbh]
\centering

    \centering
    \begin{tabular}{l@{\hspace{5pt}}c@{\hspace{5pt}}c@{\hspace{5pt}}c@{\hspace{5pt}}c@{\hspace{5pt}}c@{\hspace{5pt}}c@{\hspace{5pt}}c@{\hspace{5pt}}c@{\hspace{5pt}}c@{\hspace{5pt}}c@{\hspace{5pt}}c@{\hspace{5pt}}c}
      \toprule
      \multirow{2}{*}{\textbf{Budget}} & \multicolumn{3}{c}{\textbf{Zero-shot}} & \multicolumn{3}{c}{\textbf{Original}} & \multicolumn{3}{c}{\textbf{SPIRIT}} & \multicolumn{3}{c}{\textbf{\model}} \\
      \cmidrule(lr){2-4} \cmidrule(lr){5-7} \cmidrule(lr){8-10} \cmidrule(lr){11-13}
      & Acc $\uparrow$ & Tok $\downarrow$ & Lat $\downarrow$ & Acc $\uparrow$ & Tok $\downarrow$ & Lat $\downarrow$ & Acc $\uparrow$ & Tok $\downarrow$ & Lat $\downarrow$ & Acc $\uparrow$ & Tok $\downarrow$ & Lat $\downarrow$ \\
      \midrule
      \textbf{2K}   & 6.72  & 2021 & 0.59 & 6.34  & 1999 & 0.57 & 8.21  & 1993 & 0.56 & 13.43 & 1930 & 0.54 \\
      \textbf{4K}   & 16.79 & 3820 & 1.22 & 15.67 & 3799 & 1.20 & 20.15 & 3712 & 1.18 & 20.90 & 3594 & 1.15 \\
      \textbf{6K}   & 23.13 & 5444 & 2.07 & 22.76 & 5397 & 2.00 & 26.49 & 5237 & 1.93 & 30.60 & 4988 & 1.85 \\
      \textbf{8K}   & 25.37 & 6927 & 3.27 & 25.74 & 6882 & 3.24 & 30.60 & 6634 & 3.09 & 35.07 & 5793 & 2.38 \\
      \textbf{10K}  & 25.37 & 8336 & 5.15 & 30.97 & 8289 & 4.83 & 33.58 & 7892 & 4.62 & 36.19 & 6035 & 2.61 \\
      \textbf{12K}  & 25.75 & 9706 & 7.44 & 32.46 & 9567 & 7.10 & 34.33 & 8987 & 6.73 & 36.57 & 6128 & 2.76 \\
      \bottomrule
    \end{tabular}%

\caption{Results of different methods under different budgets on LiveCodeBench v4\_v5. We report accuracy (Acc), average number of generated tokens (Tok), and average generation latency (Lat) measured in seconds.}
\label{tab:budget_lcbv4_v5}
\end{table*}

%% file: tables/budget_lcd.tex
\begin{table*}[tbh]
    \centering
    \begin{tabular}{l@{\hspace{5pt}}c@{\hspace{5pt}}c@{\hspace{5pt}}c@{\hspace{5pt}}c@{\hspace{5pt}}c@{\hspace{5pt}}c@{\hspace{5pt}}c@{\hspace{5pt}}c@{\hspace{5pt}}c@{\hspace{5pt}}c@{\hspace{5pt}}c@{\hspace{5pt}}c}
      \toprule
      \multirow{2}{*}{\textbf{Budget}} & \multicolumn{3}{c}{\textbf{Zero-shot}} & \multicolumn{3}{c}{\textbf{Original}} & \multicolumn{3}{c}{\textbf{SPIRIT}} & \multicolumn{3}{c}{\textbf{\model}} \\
      \cmidrule(lr){2-4} \cmidrule(lr){5-7} \cmidrule(lr){8-10} \cmidrule(lr){11-13}
        & Acc $\uparrow$ & Tok $\downarrow$ & Lat $\downarrow$ & Acc $\uparrow$ & Tok $\downarrow$ & Lat $\downarrow$ & Acc $\uparrow$ & Tok $\downarrow$ & Lat $\downarrow$ & Acc $\uparrow$ & Tok $\downarrow$ & Lat $\downarrow$ \\
      \midrule
      \textbf{2K}  & 7.02  & 2028  & 0.53 & 6.14  & 2020 & 0.53 & 7.02  & 2001 & 0.53 & 10.09 & 1965 & 0.53 \\
      \textbf{4K}  & 13.16 & 3848  & 1.21 & 13.16 & 3854 & 1.21 & 16.23 & 3789 & 1.19 & 15.79 & 3758 & 1.19 \\
      \textbf{6K}  & 16.23 & 5553  & 2.04 & 16.67 & 5548 & 2.04 & 18.86 & 5407 & 2.00 & 19.30 & 5387 & 2.00 \\
      \textbf{8K}  & 19.30 & 7165  & 3.27 & 22.37 & 7104 & 3.18 & 22.37 & 6882 & 3.04 & 23.25 & 6722 & 2.88 \\
      \textbf{10K} & 19.74 & 8680  & 4.95 & 25.00 & 8485 & 4.72 & 25.00 & 8186 & 4.45 & 27.63 & 7541 & 3.48 \\
      \textbf{12K} & 21.49 & 10142 & 7.58 & 28.07 & 9717 & 7.09 & 26.32 & 9354 & 6.86 & 27.63 & 7902 & 3.83 \\
      \bottomrule
    \end{tabular}%
    \caption{Results of different methods under different budgets on LeetCodeDataset. We report accuracy (Acc), average number of generated tokens (Tok), and average generation latency (Lat) measured in seconds. }
    \label{tab:budget_lcd}%
\end{table*}

%% file: tables/budget_gsm8k.tex
\begin{table*}[tbh]
\centering

\begin{tabular}{l@{\hspace{5pt}}c@{\hspace{5pt}}c@{\hspace{5pt}}c@{\hspace{5pt}}c@{\hspace{5pt}}c@{\hspace{5pt}}c@{\hspace{5pt}}c@{\hspace{5pt}}c@{\hspace{5pt}}c@{\hspace{5pt}}c@{\hspace{5pt}}c@{\hspace{5pt}}c}
  \toprule
  \multirow{2}{*}{\textbf{Budget}} & \multicolumn{3}{c}{\textbf{Zero-shot}} & \multicolumn{3}{c}{\textbf{Original}} & \multicolumn{3}{c}{\textbf{SPIRIT}} & \multicolumn{3}{c}{\textbf{\model}} \\
  \cmidrule(lr){2-4} \cmidrule(lr){5-7} \cmidrule(lr){8-10} \cmidrule(lr){11-13}
  & Acc $\uparrow$ & Tok $\downarrow$ & Lat $\downarrow$ & Acc $\uparrow$ & Tok $\downarrow$ & Lat $\downarrow$ & Acc $\uparrow$ & Tok $\downarrow$ & Lat $\downarrow$ & Acc $\uparrow$ & Tok $\downarrow$ & Lat $\downarrow$ \\
  \midrule
  \textbf{1K} & 48.75 & 963 & 0.19 & 59.29 & 942 & 0.19 & 54.66 & 925 & 0.18 & 83.93 & 693 & 0.14 \\
  \textbf{2K} & 83.55 & 1301 & 0.27 & 86.35 & 1250 & 0.26 & 88.55 & 1118 & 0.23 & 90.75 & 753 & 0.16 \\
  \textbf{4K} & 88.65 & 1553 & 0.37 & 90.37 & 1432 & 0.34 & 90.52 & 1227 & 0.28 & 91.28 & 778 & 0.18 \\
  \textbf{6K} & 89.23 & 1714 & 0.46 & 91.05 & 1513 & 0.39 & 91.28 & 1297 & 0.33 & 91.81 & 790 & 0.20 \\
  \bottomrule
\end{tabular}%

\caption{Results of different methods under different budgets on GSM8K. We report accuracy (Acc), average number of generated tokens (Tok), and average generation latency (Lat) measured in seconds.}
\label{tab:budget_gsm8k}
\end{table*}

%% file: tables/budget_math500.tex
\begin{table*}[tbh]
\centering

\begin{tabular}{l@{\hspace{5pt}}c@{\hspace{5pt}}c@{\hspace{5pt}}c@{\hspace{5pt}}c@{\hspace{5pt}}c@{\hspace{5pt}}c@{\hspace{5pt}}c@{\hspace{5pt}}c@{\hspace{5pt}}c@{\hspace{5pt}}c@{\hspace{5pt}}c@{\hspace{5pt}}c}
  \toprule
  \multirow{2}{*}{\textbf{Budget}} & \multicolumn{3}{c}{\textbf{Zero-shot}} & \multicolumn{3}{c}{\textbf{Original}} & \multicolumn{3}{c}{\textbf{SPIRIT}} & \multicolumn{3}{c}{\textbf{\model}} \\
  \cmidrule(lr){2-4} \cmidrule(lr){5-7} \cmidrule(lr){8-10} \cmidrule(lr){11-13}
    & Acc $\uparrow$ & Tok $\downarrow$ & Lat $\downarrow$ & Acc $\uparrow$ & Tok $\downarrow$ & Lat $\downarrow$ & Acc $\uparrow$ & Tok $\downarrow$ & Lat $\downarrow$ & Acc $\uparrow$ & Tok $\downarrow$ & Lat $\downarrow$ \\
  \midrule
  \textbf{1K} & 19.00 & 1020 & 0.19 & 28.00 & 1017 & 0.19 & 18.40 & 1012 & 0.19 & 36.40 & 935 & 0.19 \\
  \textbf{2K} & 42.20 & 1804 & 0.39 & 52.00 & 1767 & 0.39 & 54.40 & 1592 & 0.36 & 59.80 & 1347 & 0.31 \\
  \textbf{4K} & 60.40 & 2629 & 0.70 & 63.80 & 2511 & 0.66 & 64.20 & 2144 & 0.57 & 70.80 & 1649 & 0.43 \\
  \textbf{6K} & 66.60 & 3100 & 0.94 & 70.60 & 2843 & 0.84 & 69.60 & 2460 & 0.74 & 71.00 & 1758 & 0.52 \\
  \bottomrule
\end{tabular}%

\caption{Results of different methods under different budgets on MATH500. We report accuracy (Acc), average number of generated tokens (Tok), and average generation latency (Lat) measured in seconds. }
\label{tab:budget_math500}
\end{table*}

%% file: tables/budget_aime24.tex
\begin{table*}[tbh]
\centering

.  \begin{tabular}{l@{\hspace{5pt}}c@{\hspace{5pt}}c@{\hspace{5pt}}c@{\hspace{5pt}}c@{\hspace{5pt}}c@{\hspace{5pt}}c@{\hspace{5pt}}c@{\hspace{5pt}}c@{\hspace{5pt}}c@{\hspace{5pt}}c@{\hspace{5pt}}c@{\hspace{5pt}}c}
   \toprule
   \multirow{2}{*}{\textbf{Budget}} & \multicolumn{3}{c}{\textbf{Zero-shot}} & \multicolumn{3}{c}{\textbf{Original}} & \multicolumn{3}{c}{\textbf{SPIRIT}} & \multicolumn{3}{c}{\textbf{\model}} \\
   \cmidrule(lr){2-4} \cmidrule(lr){5-7} \cmidrule(lr){8-10} \cmidrule(lr){11-13}
  & Acc $\uparrow$ & Tok $\downarrow$ & Lat $\downarrow$ & Acc $\uparrow$ & Tok $\downarrow$ & Lat $\downarrow$ & Acc $\uparrow$ & Tok $\downarrow$ & Lat $\downarrow$ & Acc $\uparrow$ & Tok $\downarrow$ & Lat $\downarrow$ \\
   \midrule
   \textbf{2K}   & 0.00 & 2048 & 0.93 & 16.67 & 2048 & 0.93 & 6.67 & 2048 & 0.93 & 10.00 & 1978 & 0.93 \\
   \textbf{4K}   & 20.00 & 4003 & 2.12 & 20.00 & 3984 & 2.11 & 23.33 & 3877 & 2.10 & 36.67 & 3415 & 1.99 \\
   \textbf{6K}   & 33.33 & 5682 & 3.58 & 36.67 & 5695 & 3.55 & 40.00 & 5216 & 3.32 & 36.67 & 4410 & 3.03 \\
   \textbf{8K}   & 30.00 & 7073 & 5.30 & 40.00 & 7093 & 5.14 & 46.67 & 6243 & 4.47 & 40.00 & 5159 & 4.10 \\
   \textbf{10K}  & 36.67 & 8352 & 6.76 & 46.67 & 8034 & 6.43 & 46.67 & 7198 & 5.78 & 46.67 & 5552 & 5.04 \\
   \textbf{12K}  & 40.00 & 9318 & 7.84 & 46.67 & 8990 & 7.75 & 46.67 & 8363 & 7.21 & 46.67 & 5767 & 5.88 \\
   \bottomrule
  \end{tabular}%

\caption{Results of different methods under different budgets on AIME24. We report accuracy (Acc), average number of generated tokens (Tok), and average generation latency (Lat) measured in seconds.}
\label{tab:budget_aime24}
\end{table*}

%% file: tables/budget_aime25.tex
\begin{table*}[tbh]
\centering

.  \begin{tabular}{l@{\hspace{5pt}}c@{\hspace{5pt}}c@{\hspace{5pt}}c@{\hspace{5pt}}c@{\hspace{5pt}}c@{\hspace{5pt}}c@{\hspace{5pt}}c@{\hspace{5pt}}c@{\hspace{5pt}}c@{\hspace{5pt}}c@{\hspace{5pt}}c@{\hspace{5pt}}c}
   \toprule
   \multirow{2}{*}{\textbf{Budget}} & \multicolumn{3}{c}{\textbf{Zero-shot}} & \multicolumn{3}{c}{\textbf{Original}} & \multicolumn{3}{c}{\textbf{SPIRIT}} & \multicolumn{3}{c}{\textbf{\model}} \\
   \cmidrule(lr){2-4} \cmidrule(lr){5-7} \cmidrule(lr){8-10} \cmidrule(lr){11-13}
  & Acc $\uparrow$ & Tok $\downarrow$ & Lat $\downarrow$ & Acc $\uparrow$ & Tok $\downarrow$ & Lat $\downarrow$ & Acc $\uparrow$ & Tok $\downarrow$ & Lat $\downarrow$ & Acc $\uparrow$ & Tok $\downarrow$ & Lat $\downarrow$ \\
   \midrule
   \textbf{2K}   & 6.67 & 2046 & 0.95 & 13.33 & 2048 & 0.94 & 3.33 & 2044 & 0.94 & 10.00 & 2020 & 0.95 \\
   \textbf{4K}   & 20.00 & 3851 & 2.17 & 26.67 & 3834 & 2.16 & 16.67 & 3792 & 2.15 & 20.00 & 3511 & 2.15 \\
   \textbf{6K}   & 30.00 & 5369 & 3.62 & 33.33 & 5452 & 3.63 & 36.67 & 5360 & 3.63 & 30.00 & 4484 & 3.14 \\
   \textbf{8K}   & 36.67 & 6848 & 5.30 & 36.67 & 6798 & 5.32 & 36.67 & 6611 & 5.16 & 33.33 & 5002 & 4.10 \\
   \textbf{10K}  & 40.00 & 8145 & 6.67 & 43.33 & 8026 & 6.75 & 43.33 & 7817 & 6.57 & 36.67 & 5434 & 5.10 \\
   \textbf{12K}  & 36.67 & 9442 & 8.25 & 40.00 & 9461 & 8.32 & 46.67 & 8598 & 7.78 & 36.67 & 5720 & 6.06 \\
   \bottomrule
  \end{tabular}%

\caption{Results of different methods under different budgets on AIME25. We report accuracy (Acc), average number of generated tokens (Tok), and average generation latency (Lat) measured in seconds.}
\label{tab:budget_aime25}
\end{table*}

%% file: tables/training_results_math.tex
\begin{table}[H]
\centering
\begin{tabular}{l@{\hspace{2pt}}c@{\hspace{2pt}}c}
\toprule
\textbf{Methods} & \textbf{Tokens} & \textbf{Time} \\
\midrule
Original & 5807 & 47.82 \\
\midrule
Selective Context & 3149 (-45.8\%) & 25.85 (-45.9\%) \\
LLMLingua-2 & 3478 (-40.1\%) & 28.75 (-39.9\%) \\
TokenSkip & 4728 (-18.6\%) & 39.20 (-18.0\%) \\
SPIRIT & 2858 (-50.8\%) & 23.67 (-50.5\%) \\
\midrule
\model & \textbf{1834 (-68.4\%)} & \textbf{15.36 (-67.9\%)} \\
\bottomrule
\end{tabular}

\caption{Training efficiency comparison on OpenR1-Math dataset. We report the average number of tokens per sample and training time measured in seconds per step. Percentages indicate the reduction relative to the Original baseline.}
\label{tab:training_results_math}
\end{table}